\documentclass[journal]{IEEEtai}

\usepackage{cite}
\usepackage{amsmath,amssymb,amsfonts}
\usepackage{algorithm}
\usepackage{algorithmic}
\usepackage{pgfplots}
\pgfplotsset{compat=1.18}
\usepgfplotslibrary{statistics}
\usepgfplotslibrary{groupplots}
\usetikzlibrary{patterns}
\definecolor{msteel}{RGB}{74,111,165}
\definecolor{mblue}{RGB}{47,95,140}
\definecolor{morange}{RGB}{211,151,61}
\definecolor{mteal}{RGB}{72,140,132}
\definecolor{msand}{RGB}{241,236,226}
\definecolor{mbrick}{RGB}{166,89,75}
\definecolor{mred}{RGB}{168,66,63}
\usetikzlibrary{arrows.meta, positioning, patterns}
\usepackage{graphicx}
\usepackage{subcaption}
\usepackage{textcomp}
\usepackage{xcolor}
\usepackage{url}
\usepackage{booktabs}
\usepackage{amsthm}
\newtheorem{proposition}{Proposition}
\newtheorem{theorem}{Theorem}
\newtheorem{corollary}{Corollary}
\newtheorem{definition}{Definition}
\newtheorem{remark}{Remark}
\usepackage{array}
\usepackage{listings}

\makeatletter
\def\IEEEbibitemsep{0pt plus .5pt}
\makeatother

\begin{document}

\title{Coherence by Construction: Synthesizing CTI Knowledge Graphs and the Formal Limits of Coherence Validation}

\author{ Tanzimul~Alam~Fahim,~\IEEEmembership{Student Member,~IEEE}, Mohammad~Arif~Hossain,~\IEEEmembership{Member,~IEEE}%
\thanks{Tanzimul~Alam~Fahim is with the Computational and Data Science PhD Program, MTSU, Murfreesboro, TN 37132 USA. Mohammad~Arif~Hossain is with the Department of Engineering Technology and the Quantum Research Interdisciplinary Science and Education (QRISE) Center, Middle Tennessee State University (MTSU), Murfreesboro, TN 37132 USA (e-mail: Mohammad.Hossain@mtsu.edu). }%
}

\markboth{}%
{Coherence by Construction: Synthesizing CTI Knowledge Graphs}

\maketitle

\begin{abstract}
Synthesizing relational training data such as knowledge graphs with a large
language model typically emits structured objects one at a time, which keeps
each object valid but leaves the relationships between them ungrounded. We study an alternative paradigm, \emph{generation by simulation}:
rather than describing a relational artifact, we simulate a process that
produces it and take the artifact to be its causal record. Language-model agents propose intent and a deterministic referee
resolves outcomes against an explicit world model, so a relation is emitted
only when the event it represents occurs, making coherence a property of
construction. We instantiate the paradigm for cyber threat intelligence
knowledge graphs in the STIX~2.1 standard with three engines, among them an
evolutionary engine applying language-model genetic operators to structured
graph genomes, sharing a deterministic compiler that makes malformed and
dangling output unrepresentable. We prove referential closure, causal
grounding, and identifier stability, an ablation localizes each to the referee
and world model rather than the policy, and two independent LLM judges
reproduce the rubric's engine ranking. Our central result is cautionary, and
we prove it: a validator reading any graph feature set a structure-matched
random graph can reproduce is uninformative about coherence for a
coherence-by-construction generator. Using Erd\H{o}s--R\'enyi controls we
realize the theorem four times, as structural separation, embedding transfer,
inductive link prediction, and real-data augmentation each reward a triviality
rather than coherence; we give the mechanism behind each and the conditions a valid benchmark must
meet.
\end{abstract}

\begin{IEEEImpStatement}
Synthetic data is increasingly used to train machine-learning systems where
real data is scarce, sensitive, or cannot be shared, and practitioners need
a way to confirm such data is fit for use. This work delivers a practical
caution for anyone who builds or procures synthetic relational data: for
generators that construct data by simulating a process, the checks
practitioners reach for first can certify the wrong thing; one widely
trusted check prefers random data, and another rewards sheer volume. We identify why each check fails and what sound validation requires, with
an open framework that reproduces both the generator and the analysis.
\end{IEEEImpStatement}

\begin{IEEEkeywords}
Synthetic data generation, knowledge graphs, large language models,
multi-agent systems, cyber threat intelligence.
\end{IEEEkeywords}

\section{Introduction}
\label{sec:intro}

\IEEEPARstart{M}{any} machine-learning tasks over structured, relational
data depend on a large supply of example graphs, and in many domains that
supply is the binding constraint: when data is sensitive, tied to specific
incidents, and rarely shared in trainable form, the examples that would most
help a model are the least available. Large language models make synthesis
tempting, but a relational artifact is not a bag of independently valid
objects; its value lies in the relations among them, exactly what a model
asked to describe the artifact one object at a time does not constrain. This
paper studies a different way to synthesize such data and asks how one can
tell whether the result is any good. We develop both questions in cyber
threat intelligence (CTI), which describes how attacks are carried out, the
actors, malware, infrastructure, and vulnerabilities and how they connect
into a campaign; as a knowledge graph CTI becomes machine-reasonable, and
real operational CTI is scarce for precisely the reasons above, making it
both a motivating application and a stringent test bed.

Two approaches address the shortage, each bounded in a way that leaves the
problem open. The first prompts a language model for one STIX object at a time
under schema constraints; this produces individually valid objects but leaves
the relationships unconstrained, so edges are asserted without a generative
reason. As our evaluation shows (Section~\ref{sec:results-cross}),
schema-forced generation is a competent baseline on per-graph quality; the
gap we target is not gross malformation but the absence of a causal account
of why each relationship is present, which is what a downstream model needs to
learn structure. The second extracts graphs from existing threat
reports~\cite{cheng2025ctinexus, hu2024llmtikg}, yielding coherent graphs but
only of attacks already written down, and link-prediction work over CTI
graphs reports the same shortage from the consumer
side~\cite{fieblinger2024, zhang2023edge}. Missing between them is a generator
that produces coherent CTI graphs at scale without depending on a source
corpus.

Our approach, Mirage, rests on a single change of perspective that is not
specific to security: rather than asking a language model to describe a
relational artifact, we simulate a process that would produce one and take the
resulting graph to be the record the process leaves behind. In the CTI
instantiation the process is an adversarial engagement, and a relation enters
the graph only because a corresponding action succeeded inside the simulation,
so coherence is a property of how the graph was produced rather than a
constraint applied afterward; the language model supplies tactical choices and
realistic content while a deterministic referee over an explicit world model
decides which relations hold. Agent-based simulation of cyber engagements is
not itself new; what is new is treating such a simulation as a \emph{generator
of coherent relational training data} with formal construction guarantees, and
subjecting that generator to the validation tests its intended use implies.
The second half of the paper is about those tests, where our most transferable
finding lies: the standard ways one would validate such a generator turn out
to be systematically misleading, and we explain why.

\subsection*{Contributions}
\begin{itemize}
  \item \textbf{A generation-by-simulation paradigm for coherent
        relational data.} Language-model agents propose intent and a
        deterministic referee resolves outcomes against an explicit world
        model, so the emitted graph is the causal record of a simulated
        process and coherence becomes a property of construction. We
        instantiate the paradigm for STIX~2.1 CTI graphs through three
        engines, including an evolutionary engine that, to our knowledge,
        is the first to apply language-model genetic operators to
        \emph{structured} graph genomes rather than to free text.
  \item \textbf{Construction with formal guarantees, and an ablation that
        localizes them.} A compact intermediate representation and a
        deterministic compiler make dangling and malformed output
        unrepresentable; we prove referential closure, causal grounding,
        and identifier stability, with a complexity analysis
        (Section~\ref{sec:theory}). An ablation
        (Section~\ref{sec:results-ablation}) shows the referee and world
        model author graph structure while the language-model policy
        contributes relational and semantic variety.
  \item \textbf{A theorem on the limits of coherence validation, with
        four empirical instances.} We prove that any validator whose input
        features a structure-matched random graph can match is uninformative
        about coherence for a generator that is coherent by construction
        (Theorem~\ref{thm:blind}), and that the property making such a
        generator attractive is what disarms the validator. Using
        Erd\H{o}s--R\'enyi controls, we then show that four standard
        validations, structural separation from a curated
        reference, type-level embedding transfer, properly inductive
        entity-level link prediction, and augmentation of real training
        data, are each uninformative for this class of generator: the
        inductive test favours a random-graph corpus, and augmentation
        gains are matched by a size-matched random corpus; for each we
        identify the mechanism and the conditions a valid benchmark must
        meet. We regard this, rather than the generator alone, as the
        paper's most transferable contribution.
\end{itemize}

These contributions compose established components rather than introducing a
new learning algorithm; the contribution is in their arrangement under a
deterministic referee and compiler and in the evaluation analysis that
follows.

\section{Related Work}
\label{sec:related}

\subsection{Synthetic CTI Generation and CTI Knowledge Graphs}
\label{sec:rw-gen}
Using language models to synthesize training data is now widespread across
domains~\cite{long2024syntheticsurvey}, but most recent work that builds CTI
knowledge graphs with language models is
\emph{extractive}, converting existing unstructured reports into structured
graphs: CTINexus uses optimized in-context learning~\cite{cheng2025ctinexus},
LLM-TIKG fine-tunes a smaller model for entity, relation, and TTP
extraction~\cite{hu2024llmtikg}, CTI-Thinker adds LoRA and retrieval-augmented
reasoning over an ATT\&CK-aligned graph~\cite{ctithinker2026}, and a
related line infers attack-technique hypotheses over a fused knowledge
base~\cite{kaiser2023}. All share one precondition, a source corpus, so their
output is bounded by what has already been written down; Mirage runs in the
opposite direction, producing a STIX graph from a simulated engagement with no
source report.

\subsection{Autonomous Cyber Agents and Cyber-Range Simulation}
\label{sec:rw-sim}
The closest body of work simulates adversarial engagements with multiple
agents, but to train and evaluate the agents rather than to emit data. A
substantial literature applies reinforcement learning to autonomous cyber
attack and defense, training red and blue agents in simulated or emulated
networks and addressing the simulation-to-emulation fidelity such transfer
requires~\cite{nguyen2023drlcyber}; the most directly comparable system embeds
deep-RL attacker and ML defender agents in a cyber range, validated against
CICIDS2017 and UNSW-NB15~\cite{agrawal2026masim}. A separate strand uses
language models to generate multi-stage attacks directly and reports that their
unconstrained use is unreliable~\cite{sen2024smartgrid}. Mirage differs on four
axes (detailed in the supplementary material): it generates data rather than
training agents, outputs a STIX~2.1 graph rather than a policy, uses
language-model rather than reinforcement-learning agents, and lets a
deterministic referee over an explicit world model, not the agents, decide
which relationships enter the graph. That last axis is the most consequential,
and the smart-grid result is what makes it necessary: the consistency an
unconstrained language model fails to maintain is supplied here by the world
model. Our ``first to'' claims are scoped to this combination, which to our
knowledge no prior system targets.

\subsection{LLM Multi-Agent Adversarial Games and Mode Collapse}
\label{sec:rw-llm}
A growing literature pits language models against each other in red- and
blue-team roles, but most of it targets the safety of the language model
itself~\cite{evolvingredteam2024, rethinkingredblue2025}; the mode-collapse
obstacle identified there applies to our setting and motivates the evolutionary
engine of Section~\ref{sec:method-evo}, where we treat corpus-level diversity
as a design goal. Here, by contrast, the agents act against a simulated network
and the model is the policy rather than the victim. The extended discussion is
in the supplementary material.

\subsection{KG Embedding for CTI and Evolutionary Data Generation}
\label{sec:rw-kge}
The downstream consumers of CTI knowledge graphs motivate this work and report
being short of data~\cite{fieblinger2024, zhang2023edge, peng2020lineare}. A
parallel line learns graph distributions directly with deep generative
models~\cite{guo2022graphgen}; Mirage differs in kind, since it does not fit a
distribution but emits each graph as the causal record of a simulated process,
so validity is a construction guarantee rather than a learned approximation.
The evolutionary engine has a separate lineage: using a
language model as a genetic operator with deterministic selection is
established for text and code~\cite{geneticprompt2025},
but in every such system the genome is linguistic. Engine~C applies the
paradigm to a graph genome that compiles to STIX, which to our knowledge is the
first use of language-model-driven evolutionary operators on structured
threat-intelligence knowledge graphs. The extended discussion is in the
supplementary material.

\section{Background and Threat Model}
\label{sec:background}

\subsection{STIX 2.1 and CTI Knowledge Graphs}
Structured Threat Information Expression (STIX) 2.1 is the OASIS standard for
machine-readable cyber threat intelligence~\cite{stix21spec}. It models
intelligence as a graph of typed objects: \emph{STIX Domain Objects} (SDOs)
represent entities such as threat actors, malware, attack patterns,
vulnerabilities, and infrastructure, and \emph{STIX Relationship Objects}
(SROs) connect them with typed, directed relationships, such as a threat actor
\textsf{uses} a piece of malware; a \emph{bundle} forms a knowledge graph with
SDOs as nodes and SROs as edges. The analytical value lies in the connectivity
among objects, since the pattern of relationships encodes how an attack
unfolds~\cite{sikos2023survey}, and CTI knowledge graphs are increasingly the
substrate for automated threat analysis and mining~\cite{sun2023ctimining,
gao2022hincti}.

\subsection{Knowledge Graph Embedding as the Downstream Task}
A primary consumer of CTI knowledge graphs is knowledge graph embedding (KGE),
which learns vector representations of nodes and relationships for link
prediction~\cite{wang2017kgesurvey, ji2022kgsurvey}. A broad family of scoring
models exists, and their relative performance is benchmark-dependent rather
than universally ordered~\cite{rossi2021kgecomparative}. These methods are
data-hungry, so the scarcity of shareable CTI graph data is the gap a generator
is meant to fill; Mirage's output is directly consumable by standard KGE
pipelines.

\subsection{Scope and Threat Model}
\label{sec:scope}
Mirage's wargaming engine models an enterprise network intrusion: an attacker
moving through a network of hosts toward a high-value objective, against a
defender that monitors for and responds to suspicious activity. The world model
represents hosts, their services and vulnerabilities, exploitation under
uncertainty, lateral movement, and alert-based detection; this suffices to
generate coherent multi-stage engagement graphs and is not a complete emulation
of a real enterprise environment. We claim these graphs are structurally
coherent with relationships causally grounded in a simulated engagement, but
not that they are behaviorally or statistically indistinguishable from real
operational intelligence; we lack an operational reference against which such a
claim could be tested, and Section~\ref{sec:results-finding2} shows the most
natural candidate, MITRE ATT\&CK, cannot stand in for one. ``Suitable as data
for downstream machine learning'' is therefore a claim about structural
properties, not operational fidelity.

\section{Methodology}
\label{sec:method}

\subsection{Overview: From Schema Enforcement to Causal Generation}
\label{sec:method-overview}
\begin{figure*}[!t]
\centering
\includegraphics[width=0.95\textwidth]{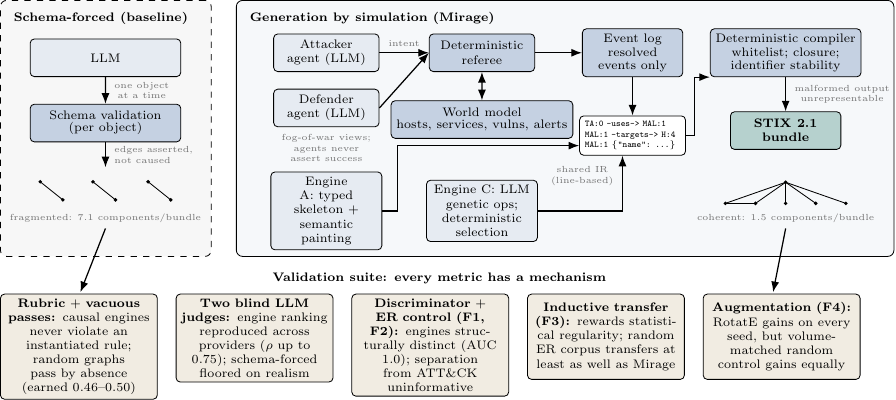}
\caption{The two generation paradigms and the validation suite: schema-forced per-object generation (left) versus generation by simulation in Mirage (right); Section~\ref{sec:method} details the data flow.}
\label{fig:architecture}
\end{figure*}

The conventional way to generate synthetic STIX data prompts a language model
for one object at a time and validates each against the schema, which keeps
every object well-formed but leaves the relationships, where a
threat-intelligence graph's value lies, asserted without any underlying reason.
Mirage inverts this: we \emph{simulate} an adversarial engagement and treat
the resulting graph as the record it leaves behind, so a relationship appears
only because a corresponding action succeeded inside the simulation. The
language model supplies realistic names, descriptions, and tactical choices but
never decides which relationships hold; a deterministic referee over an
explicit world model does. Mirage exposes three engines sharing a common output
path: Engine~A builds topology procedurally and uses the model only for
semantics; Engine~B, the central contribution, generates a graph as the
by-product of a turn-based attacker--defender simulation; and Engine~C evolves
a population using the model as a genetic operator. We call Engines~B and~C the
\emph{causal engines}, since in both a relationship enters the graph only as
the consequence of a resolved event. All three emit the same intermediate
representation, which one deterministic compiler turns into a valid STIX~2.1
bundle; we describe it first, as it is the contract every engine writes to.
Figure~\ref{fig:architecture} shows the overall data flow.

\begin{figure}[tbp]
\centering
\includegraphics[width=0.8\columnwidth]{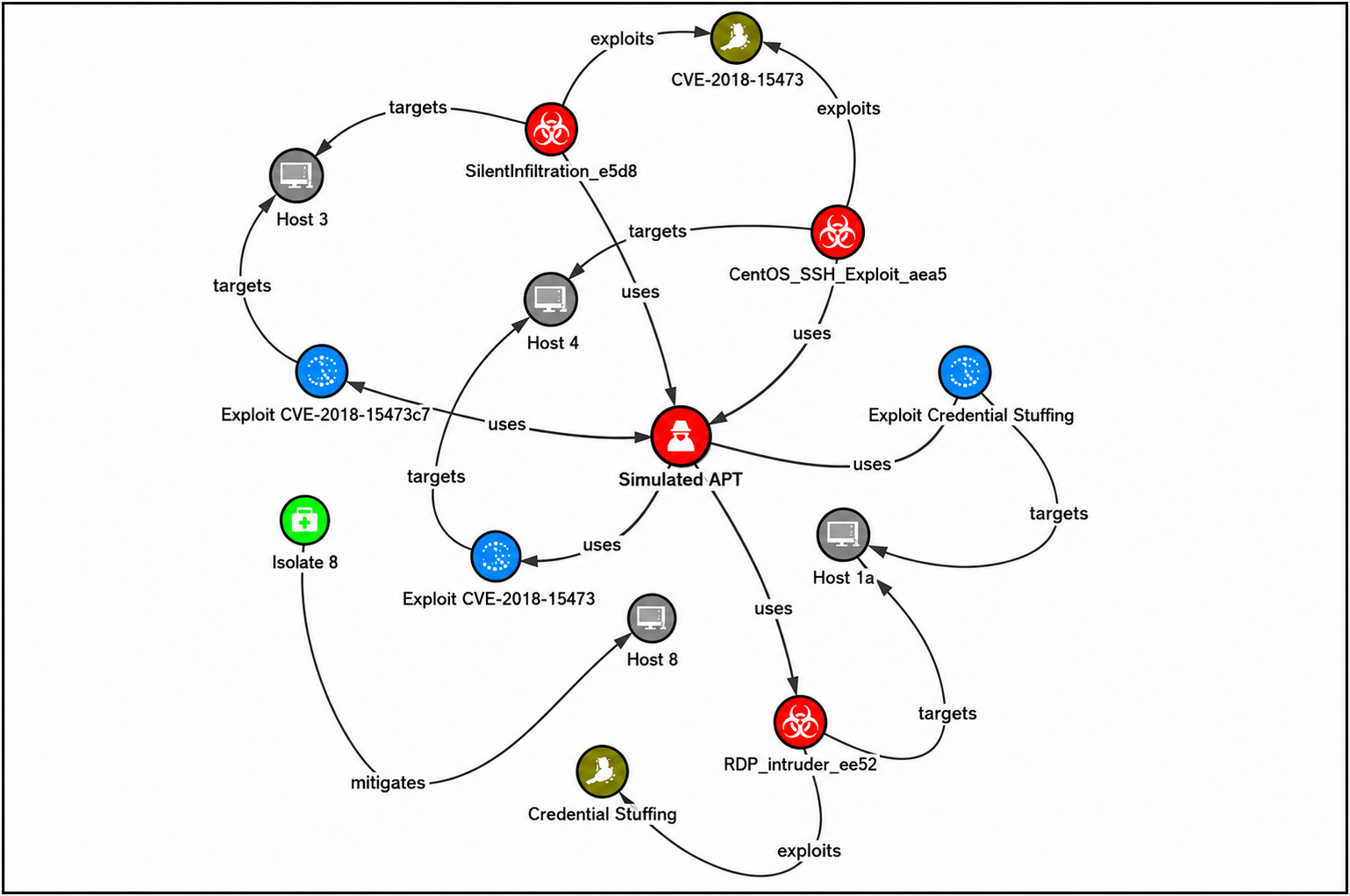}
\caption{An example of STIX~2.1 graph from the wargaming engine (OASIS STIX Visualizer): a threat actor \textsf{uses} malware and attack patterns, malware \textsf{targets} hosts and \textsf{exploits} vulnerabilities, and a course-of-action records one turn's mitigation.}
\label{fig:example-graph}
\end{figure}

\subsection{Intermediate Representation and Deterministic Compilation}
\label{sec:method-ir}

STIX~2.1 is a deeply nested JSON format; asking a generator to emit it
directly forces it to manage UUID allocation, cross-references, and required
fields while making content decisions, which is where dangling references and
malformed objects enter. Every engine instead targets a small line-based
intermediate representation (IR), and translation from IR to STIX is the sole
responsibility of a deterministic compiler. The IR describes a graph one
statement per line: a relationship statement names a typed source, a verb,
and a typed target, and a property statement attaches attributes to a named
node, with local human-readable identifiers:

\begin{lstlisting}[basicstyle=\ttfamily\footnotesize,frame=single,
                   xleftmargin=1em,framexleftmargin=1em,
                   breaklines=true,breakatwhitespace=true,
                   postbreak=\mbox{\textcolor{gray}{$\hookrightarrow$}\space},
                   columns=fullflexible]
TA:0 -uses-> MAL:1
MAL:1 -communicates-with-> INFRA:2
MAL:1 {"name": "DarkPulse", "is_family": false}
\end{lstlisting}

\noindent
Compilation enforces two properties by construction. \emph{Identifier
stability}: each distinct IR identifier is minted to exactly one STIX UUID
within a compilation, while a fresh compilation draws fresh UUIDs so
separately generated graphs never collide. \emph{Referential closure}: every
node referenced by a relationship is created during the same pass that
resolves it, so dangling references are unrepresentable rather than detected
and repaired. Required STIX fields absent from the IR are supplied with
schema-conformant defaults, and the relationship vocabulary is enforced here:
a verb is admitted only if the (source type, verb, target type) triple
appears in a whitelist derived from the STIX~2.1
specification~\cite{stix21spec}, and a violating statement is logged and
skipped rather than silently dropped, so a malformed line stays visible in
the build record. Identifier stability and referential closure are stated
formally as Propositions~\ref{prop:stability} and~\ref{prop:closure} in
Section~\ref{sec:theory}. Both invariants are stated over the endpoints of
emitted SROs and do not extend to embedded reference fields
(\texttt{object\_refs} of \textsf{report}, \textsf{grouping}, \textsf{note},
\textsf{opinion}, or the \textsf{sighting} SRO), which the binary-triple IR
does not express and which compile as opaque strings; extending closure to
them is a planned addition. Because a whitelist-violating statement is skipped
rather than aborting the build, a connecting edge can be dropped at compile
time, and the resulting component structure is what
Section~\ref{sec:results-struct-decomp} measures.

\subsection{Engine A: Procedural Topology with Semantic Painting}
\label{sec:method-proc}
Engine~A separates \emph{what shape} a threat graph has from \emph{what its
nodes mean}. A directed scale-free topology is generated with the
Barab\'asi--Albert preferential-attachment model~\cite{barabasi1999},
producing the heavy-tailed degree distribution of real attack graphs; node
types are assigned by connectivity rather than at random, since edge
direction carries meaning in STIX (high-out-degree nodes initiate
relationships, high-in-degree nodes are targets), with self-loops and
parallel edges removed first. Only once the typed skeleton exists does the
model enter, with a narrow job: given a node's type and position it supplies
realistic content (a malware name, a concrete CVE, an infrastructure label).
This ``semantic painting'' never adds, removes, or redirects an edge; it
annotates a fixed structure in batches, at a cost that scales with nodes
rather than relationships.

\subsection{Engine B: Multi-Agent Wargaming}
\label{sec:method-war}

Engine~B is the central contribution and the clearest realization of the
generate-by-simulation principle: rather than describing an attack, it
\emph{plays} one. An attacker and defender take turns acting on a network
whose ground truth is fixed in advance, and the STIX graph is assembled from
the actions that actually succeed, a faithful record of a single engagement
(Fig.~\ref{fig:example-graph} shows a representative output).

\subsubsection{World model and fog of war}
A scenario samples a network of hosts, each with running services and,
independently, vulnerabilities; a host is exploitable through a service only
if it carries a matching vulnerability, and this ground truth is hidden from
both agents. The attacker begins at a single entry host and perceives the
network only through its actions: a scan reveals a reachable host's services
and OS but not whether they are vulnerable, forcing exploitation under
uncertainty and making a failed exploit informative to both attacker and
defender.

\subsubsection{Agents propose intent; a referee resolves outcomes}
On each turn the attacker and defender are each asked for an \emph{intent}: a
choice of action and target drawn from their current view (``scan host $h$,''
``exploit service $s$ on host $h$,'' or for the defender ``isolate host
$h$''). The agents never assert success and never name a STIX relationship.
Whether an action succeeds is decided by a deterministic referee consulting
the hidden ground truth: an exploit against a vulnerable service succeeds with
probability $0.85$ and against a patched service with probability $0.05$
(calibrated parameters in the supplementary material), and only on success
does the attacker gain control of the host and the corresponding STIX edges
get emitted. The referee thus decides which relationships are emitted while
the policy shapes the distribution of attempted actions: hallucinated
relationships are impossible by construction, since no edge is emitted without
a resolved event, while the topology remains partly a function of the policy
model (Section~\ref{sec:discussion}). Algorithm~\ref{alg:turn} gives the
single-turn procedure.

\begin{algorithm}[!t]
\caption{One turn of the wargaming engine. The language model supplies only the \emph{intents} (lines~2--3); the deterministic referee decides every outcome and every emitted edge (lines~4--16).}
\label{alg:turn}
\small
\begin{algorithmic}[1]
\STATE \textbf{Input:} world state $W$ (hidden ground truth); attacker view
       $V_A$; defender view $V_D$; alert map $\alpha$; emitted edge set $E$
\STATE $a_{\mathrm{red}} \leftarrow \textsc{AttackerIntent}(V_A)$
       \COMMENT{LLM: action $+$ target, under fog of war}
\STATE $a_{\mathrm{blue}} \leftarrow \textsc{DefenderIntent}(V_D, \alpha)$
       \COMMENT{LLM: isolate / deploy / idle}
\STATE $r \leftarrow \textsc{Resolve}(a_{\mathrm{red}}, W)$
       \COMMENT{referee consults ground truth}
\IF{$a_{\mathrm{red}}$ is \texttt{exploit} on host $h$ via service $s$}
    \STATE $p \leftarrow 0.85$ if $s$ is vulnerable on $h$ else $0.05$
    \IF{$\textsc{rand}() \le p$ \textbf{and} $h$ not already owned}
        \STATE mark $h$ compromised in $W$
        \STATE $E \leftarrow E \cup \{\textsc{ta}\!\to\!\textsc{mal},
               \textsc{mal}\!\to\!h, \textsc{mal}\!\to\!\textsc{cve},
               \textsc{ta}\!\to\!\textsc{ap}, \textsc{ap}\!\to\!h\}$
               \COMMENT{kill-chain edges emitted only on success}
    \ENDIF
    \STATE $\alpha[h] \leftarrow \alpha[h] + 1$
           \COMMENT{success or failure both raise an alert}
\ELSIF{$a_{\mathrm{red}}$ is \texttt{scan} from foothold $f$}
    \STATE reveal neighbours' services/OS in $V_A$ (never vulnerabilities)
    \STATE $\alpha[f] \leftarrow \alpha[f] + 1$
\ENDIF
\IF{$a_{\mathrm{blue}}$ is \texttt{isolate}~$h$ \textbf{and}
    $\alpha[h] \ge \textsc{AlertThreshold}$}
    \STATE remove $h$ from attacker reach;
           $E \leftarrow E \cup \{\textsc{coa}\!\to\!h\}$
\ELSIF{$a_{\mathrm{blue}}$ is \texttt{deploy} on malware $m$}
    \STATE $E \leftarrow E \cup \{\textsc{ind}\!\to\!m\}$
\ENDIF
\STATE \textbf{terminate if:} a crown jewel is compromised (attacker wins),
       all footholds isolated and stalled (eviction), or the turn budget
       $\max(25,\lfloor 1.5n\rfloor)$ is reached
\end{algorithmic}
\end{algorithm}

\subsubsection{Detection, eviction, and termination}
The defender does not see the attacker directly; it accumulates alerts from
failed exploits, successful compromises, and scans from compromised footholds,
and when a host's alerts reach a fixed threshold it may isolate the host. The
alert weighting is a calibrated parameter (Section~\ref{sec:discussion}). An
engagement ends when the attacker reaches a designated crown-jewel host, the
defender evicts the attacker from every foothold, or a turn budget of
$\max(25, \lfloor 1.5\,n \rfloor)$ turns for $n$ hosts is exhausted; the crown
jewel is chosen among hosts reachable from the entry point along directed
edges, and the entry host must have outbound connectivity for lateral movement.
Graph size and shape then \emph{emerge} from how the engagement plays out
rather than from a parameter, tuned to a low-but-nonzero defender-win target
(detailed mechanics and parameters in the supplementary material).

\subsection{Engine C: Evolutionary Optimization over STIX Graphs}
\label{sec:method-evo}
Engine~C addresses mode collapse, the tendency of a language-model generator
to keep producing variations on a narrow set of templates so the corpus
carries little real diversity~\cite{evolvingredteam2024}. It treats a set of
generated graphs as a population and applies an evolutionary loop selecting
for quality while rewarding distance from the rest of the population. The
population is seeded from the wargaming and procedural engines; each
generation is produced by mutation and crossover, with the model as a genetic
operator \emph{only}: given one or two parent graphs in the IR it proposes an
offspring, for example by substituting a technique or splicing one kill chain
into another. The model never scores candidates and never decides what
survives: selection is deterministic, driven by a fitness combining quality
(Section~\ref{sec:eval}) and novelty in a fixed, untuned $0.8/0.2$ weighting;
the ablation of Section~\ref{sec:results-ablation} reports the effect of
removing the novelty term. Two safeguards prevent reward-hacking: every
candidate must clear a structural sanity floor before selection, and novelty
is normalized across the whole population rather than within a lineage. The
methodological contribution is the application of a language-model genetic
operator to a graph genome that compiles to STIX, rather than to free text or
code~\cite{geneticprompt2025}; to our knowledge this is
the first such application to structured threat-intelligence knowledge graphs.
We are clear about what we do not establish: the intended benefit is
corpus-level diversity, which we do not measure here, and under our limited
evolutionary budget per-graph relational diversity does not exceed that of the
wargaming engine (Section~\ref{sec:results-cross}). We present Engine~C as a
demonstration that the genetic-operator paradigm transfers to structured
graphs, with a direct corpus-level diversity measurement as future work.

\section{Theoretical Analysis}
\label{sec:theory}

The guarantees claimed in Section~\ref{sec:method} are properties of the
compiler and the referee, not of the language model, so they can be stated
and proved without reference to model behaviour; extended remarks and the
per-engine cost derivations are in the supplementary material.

\subsection{Compiler Correctness}
\label{sec:theory-compiler}

We model the compiler as the function $\mathsf{compile}: \mathrm{IR} \rightarrow
\mathrm{Bundle}$ realized by \texttt{compile\_dsl}. Its input is an IR document
that parses (Section~\ref{sec:method-ir}) to a finite set of relationship
statements $T$ and property statements $P$, where each
$t = (\sigma_s, \iota_s, \rho, \sigma_t, \iota_t) \in T$ names a source type and
local identifier, a relationship verb, and a target type and local identifier.
The \emph{node key} of an endpoint is the pair $(\sigma, \iota)$, and the
\emph{node universe} $\mathcal{N}$ is the set of node keys occurring in
$T \cup P$. The compiler proceeds in two phases: a \emph{minting phase} assigns
every node key in $\mathcal{N}$ a fresh STIX UUID, recorded in a map
$\mu:\mathcal{N}\rightarrow\mathrm{UUID}$, and an \emph{emission phase} creates
one STIX Domain Object per entry of $\mu$ and then, for each $t\in T$ whose
type-verb-type signature is admitted by the whitelist $\mathcal{W}$, one STIX
Relationship Object whose \texttt{source\_ref} and \texttt{target\_ref} are
built from $\mu(\sigma_s,\iota_s)$ and $\mu(\sigma_t,\iota_t)$.

\begin{proposition}[Referential closure]
\label{prop:closure}
For every IR document and every STIX Relationship Object $r$ that
$\mathsf{compile}$ emits, both endpoints of $r$ reference STIX Domain Objects
that $\mathsf{compile}$ also emits in the same bundle. Equivalently, the
compiler cannot emit a relationship with a dangling \texttt{source\_ref} or
\texttt{target\_ref}.
\end{proposition}

\noindent\emph{Proof (supplementary material).} Both endpoints of any emitted
relationship are minted into $\mathrm{dom}(\mu)$ before the emission phase
creates one object per key, and the whitelist can only remove relationships,
never add a dangling one.\hfill$\square$

\begin{remark}
Proposition~\ref{prop:closure} is stated over emitted SRO endpoints: it does
not assert that every relationship \emph{proposed} in the IR is emitted (a
statement whose signature is not in $\mathcal{W}$ is skipped and logged), nor
does it govern embedded reference fields, which lie outside the binary-triple
IR. The relationships the compiler does emit are never dangling, the property
direct per-object LLM generation fails to provide.
\end{remark}

\begin{proposition}[Identifier stability]
\label{prop:stability}
Within a single compilation, every reference to a given node key resolves to one
and the same STIX UUID, and two separate compilations assign UUIDs that coincide
only with probability negligible in the UUID length.
\end{proposition}

\noindent\emph{Proof (supplementary material).} The guarded map $\mu$ assigns
each key a UUID once, giving intra-compilation stability; across compilations,
a union bound over $122$-bit version-4 UUIDs bounds any collision by
$|\mathcal{N}_1|\,|\mathcal{N}_2|\,2^{-122}$.\hfill$\square$

\subsection{Causal Grounding of the Wargaming Referee}
\label{sec:theory-referee}

The wargaming engine assembles its IR from emitted relationship statements
produced during simulation. We show this list contains a statement only as the
consequence of a resolved event, the sense in which the graph is the causal
record of the engagement.

\begin{definition}[Resolved event]
\label{def:resolved-event}
A \emph{resolved event} in a turn is one of: (i) a successful exploit, namely an
attacker exploit action against an open service of a host for which the referee's
probabilistic resolution against the hidden ground truth returns success; (ii) a
successful defender indicator deployment; or (iii) a successful host isolation,
namely a defender isolation action against a host whose accumulated alert count
has reached \textsc{AlertThreshold}.
\end{definition}

\begin{proposition}[Causal grounding]
\label{prop:causal}
Every relationship statement appended to the wargaming engine's emission list,
and hence every SRO the compiler later emits for a wargaming graph, corresponds
to a resolved event of the engagement. No relationship statement is appended on a
turn whose action does not resolve to one of the events in
Definition~\ref{def:resolved-event}.
\end{proposition}

\begin{proof}
The emission list is written only by the referee routine \textsc{Resolve}, at a
fixed set of program points; by inspection there are exactly seven. Five form
the exploit edge bundle: the threat actor uses a malware and an attack pattern,
the malware targets the host infrastructure and exploits the vulnerability, and
the attack pattern targets the host. All five lie on the branch guarded by the
successful-resolution test, so they execute only when the probabilistic
resolution of an exploit against an actually open service returns success; the
failed-exploit and refused-connection branches return without writing. The
sixth point lies on the defender indicator-deployment action, and the seventh
on the host-isolation action guarded by the alert-threshold test. Each of the
three groups coincides with one resolved-event type of
Definition~\ref{def:resolved-event}, and no other program point writes to the
emission list, so every appended statement corresponds to a resolved event;
because the compiler emits SROs only from this list, the conclusion carries to
the bundle. Edges introduced by the topology sampler that are never traversed
by a resolved action are not written to the emission list and never become
SROs.
\end{proof}

\begin{remark}[Internal consistency versus external realism]
\label{rem:internal-external}
Proposition~\ref{prop:causal} is a statement of \emph{internal} consistency:
every edge is the trace of an event in the world model. It does not assert
\emph{external} realism, that the distribution of such traces matches that of
real operational intrusions. These are distinct properties; the gap between them
is what Section~\ref{sec:results} probes and Section~\ref{sec:discussion}
identifies as the central open problem, and the causal-grounding guarantee
must not be read as a realism guarantee. The referee maintains two invariants
used in the proof: a host's compromised flag is set only on a successful
exploit and never cleared except by isolation, and its alert counter is
monotonically nondecreasing within an engagement, both immediate by inspection
of the only writes to these fields.
\end{remark}

\subsection{Complexity}
\label{sec:theory-complexity}

We give the asymptotic cost of each stage in terms of the IR size and the
simulation parameters. Let $m$ be the number of relationship statements and $p$
the number of property statements in an IR document, and let $n=|\mathcal{N}|$ be
the number of distinct node keys, so that $n \le 2m + p$.

\begin{proposition}[Compilation complexity]
\label{prop:compile-complexity}
$\mathsf{compile}$ runs in $O(m + p)$ expected time and $O(n)$ space, treating
hash-map operations and whitelist membership as expected-constant time.
\end{proposition}

\begin{proof}
The minting phase performs a constant number of hash-map insertions and lookups
per statement, hence $O(m+p)$ expected time, and stores one UUID and one
property record per distinct key, hence $O(n)$ space. The emission phase
creates one object per key, $O(n)$, and processes each relationship statement
once with a constant number of whitelist and map operations, $O(m)$. Summing
gives the bound.
\end{proof}

For the wargaming engine with $H$ hosts and turn budget
$B=\max(25,\lfloor 1.5H\rfloor)$, each turn issues a
constant number of language-model calls and $O(H)$ worst-case referee work, so
the engine runs in $O(B\cdot H)=O(H^2)$ referee time exclusive of model latency
and emits $O(B)$ relationship statements. Engine~A is dominated by the
scale-free construction at $O(n\log n)$ plus one batched model call, and
Engine~C over $G$ generations with population size $\pi$ costs $O(G\pi)$
operator calls plus $O(G\pi^2\bar{s})$ for global novelty at mean graph size
$\bar{s}$; the derivations are given in the supplementary material.
Language-model latency dominates wall-clock time in practice, as
Section~\ref{sec:eval-setup} reports.

\subsection{Graph-Theoretic Characterization and the Evolutionary Loop}
\label{sec:theory-graph}

Two structural properties follow from the generation process rather than from
measurement. For Engine~A the node-degree distribution is that of the
underlying directed scale-free construction (up to removal of self-loops and
parallel edges), hence heavy-tailed by construction. For the wargaming
engine the emitted edge multiset is a union of fixed-size bundles, one per
resolved compromise, plus at most one indicator edge per deployment and one
mitigation edge per isolation, so emitted edges are an affine function of the
three resolved-event counts and mean degree is governed by bundle size
rather than host count; this is why graph size tracks
engagement length rather than network size
(Section~\ref{sec:results-robustness}). For Engine~C we state a guard property
rather than a convergence guarantee: selection admits a
candidate only if it clears the structural sanity floor
(Section~\ref{sec:method-evo}), so every generation consists of structurally
valid graphs regardless of what the operator proposes; we claim no
diversity-preservation theorem and treat convergence as open. The
supplementary material expands both, including the analytic
parameter-to-structure map we identify as the natural completion.

\subsection{The Limits of Coherence Validation}
\label{sec:theory-eval}

The guarantees above secure internal consistency by construction. We now ask
what that construction implies for \emph{evaluation}, and prove that the
property which makes the generator attractive is the same property that
disarms a broad class of validators, the formal counterpart of the empirical
Findings~2 through~4. Fix a finite property space and let a \emph{coherence
property} $\Phi$ map a graph to $\{0,1\}$. A generator $\mathcal{G}$ is
\emph{coherent by construction} for $\Phi$ if
$\Pr_{G\sim\mathcal{G}}[\Phi(G)=1]=1$, and a \emph{validator} is any statistic
$V$ mapping a graph to a real score intended to be larger when the graph is
more coherent. The question is whether high $V$ on $\mathcal{G}$'s output is
evidence that $\mathcal{G}$ produces coherence or evidence about something
else $V$ happens to measure.

\begin{definition}[Construction-blind validator]
\label{def:blind}
A validator $V$ is \emph{construction-blind} for $\Phi$ on a generator pair
$(\mathcal{G},\mathcal{G}_0)$ if its score distribution is identical on both
generators, $V_\#\mathcal{G}=V_\#\mathcal{G}_0$ (equality of pushforward
laws), where $\mathcal{G}$ is coherent by construction for $\Phi$ and
$\mathcal{G}_0$ is a null generator with $\Pr_{G\sim\mathcal{G}_0}[\Phi(G)=1]=0$.
\end{definition}

A construction-blind validator scores coherent and incoherent generators
identically, so no threshold distinguishes them and its endorsement of
$\mathcal{G}$ is uninformative about $\Phi$. The following theorem gives a
sufficient condition, met by the validators in our study, under which this
holds necessarily.

\begin{theorem}[Coherence-blindness of structure-matched validators]
\label{thm:blind}
Let $\mathcal{G}$ be coherent by construction for a coherence property
$\Phi$, and let $V=f\circ\psi$ factor through a feature map $\psi$ into a
score, where $\psi$ depends on a graph only through a feature vector that
omits $\Phi$ (formally, $\psi(G)=\psi(G')$ whenever $G,G'$ agree on the
features $\psi$ reads, even if $\Phi(G)\neq\Phi(G')$). If there exists a
null generator $\mathcal{G}_0$ with $\Pr_{\mathcal{G}_0}[\Phi=1]=0$ whose
induced feature law matches that of $\mathcal{G}$, that is
$\psi_\#\mathcal{G}_0=\psi_\#\mathcal{G}$, then $V$ is construction-blind for
$\Phi$ on $(\mathcal{G},\mathcal{G}_0)$.
\end{theorem}

\begin{proof}
$V=f\circ\psi$ depends on $G$ only through $\psi(G)$, so the law of $V$ under
any generator is the pushforward of that generator's feature law through
$f$: $V_\#\mathcal{G}=f_\#(\psi_\#\mathcal{G})$ and likewise for
$\mathcal{G}_0$. By hypothesis $\psi_\#\mathcal{G}_0=\psi_\#\mathcal{G}$,
hence $f_\#(\psi_\#\mathcal{G}_0)=f_\#(\psi_\#\mathcal{G})$ and therefore
$V_\#\mathcal{G}_0=V_\#\mathcal{G}$. Since $\mathcal{G}_0$ realizes
$\Phi=0$ almost surely while $\mathcal{G}$ realizes $\Phi=1$ almost surely,
$V$ satisfies Definition~\ref{def:blind}.
\end{proof}

The content of the theorem is the antecedent: a validator is disarmed
exactly when an incoherent null can match the features it reads. Coherence
by construction is what makes such a null easy to build, because the
generator's coherence is invisible to any feature set that does not encode
$\Phi$ itself, and a structure-matched random graph then reproduces the
validator's input distribution while possessing none of the coherence. Two
consequences make the abstract statement concrete.

\begin{corollary}[The discriminator and the random-graph control]
\label{cor:disc}
A two-sample or classifier-based realism test reads only the graph-level
feature vector $\psi$ of Section~\ref{sec:eval} and is therefore of the form
$V=f\circ\psi$. Taking $\mathcal{G}_0$ to be the Erd\H{o}s--R\'enyi control
of Finding~2, matched to $\mathcal{G}$ on those features, the test is
construction-blind for relational coherence: it cannot separate Mirage from
a feature-matched random graph on the basis of coherence, which is the
formal reason the size-matched MMD comparison, not the raw discriminator
AUC, is the only informative reading.
\end{corollary}

\begin{corollary}[Embedding transfer and augmentation]
\label{cor:transfer}
A knowledge-graph-embedding transfer or augmentation score is a function of
the multiset of training triples through the learned embedding, a feature
map that reads relational \emph{statistics} (degree, relation frequency,
neighbourhood overlap) and not per-scenario causal coherence. A
volume-matched random corpus that matches those statistics therefore induces
the same score law in the limit of the theorem, which is the mechanism
behind Findings~3 and~4: the inductive test favours a random corpus, and the
augmentation gain is reproduced by a size-matched random control.
\end{corollary}

The theorem reframes the paper's negative results as instances of one
positive fact and carries a constructive contrapositive: a validator escapes
construction-blindness only if its feature map reads a correlate of $\Phi$ no
structure-matched null can reproduce. This is the design constraint
Section~\ref{sec:discussion} places on a valid benchmark, now derived rather
than asserted; the LLM judge of Section~\ref{sec:results-judge} is one
attempt to read such a correlate, its imperfect but provider-independent
agreement with the rubric consistent with a feature map that partially
encodes $\Phi$.

\section{Implementation}
\label{sec:impl}
Mirage is implemented in Python and released as open
source.\footnote{Source code and documentation:
\url{https://anonymous.4open.science/r/Project-Mirage/}.} Every engine writes
the IR of Section~\ref{sec:method-ir} and a single compiler module turns it
into a STIX~2.1 bundle through one canonical object-builder, so the closure and
default-population guarantees hold uniformly; every bundle is validated by the
official \texttt{stix2} parser, and Engine~A topology uses
NetworkX~\cite{networkx2008}. The policy and operator model throughout is
OpenAI's GPT-4o mini via the public API; this is part of the method, since
tactical decisions, semantic painting, and operator proposals all depend on
model behaviour, though in every case the model proposes and a deterministic
component decides. We pin the dated snapshot, temperature, and per-role seeds
and release the evaluation records, generated bundles, and result files behind
every table, so reported numbers are verifiable without re-querying the model.

\section{Evaluation}
\label{sec:eval}

A generator of synthetic data is useful only if its output can be told apart
from noise for the right reasons, so our evaluation is built around a
discriminator and the controls that make its verdict interpretable.

\subsection{Metrics and Protocol}
\label{sec:eval-protocol}

Each generated graph is scored on four axes: \emph{compliance} (the graph
parses as valid STIX~2.1, checked with the official \texttt{stix2} parser),
\emph{structural} quality (the graph's shape is consistent with a coherent
attack graph), \emph{semantic} quality (node content and relationships obey
domain rules), and \emph{diversity} (the Shannon entropy of the
relationship-type distribution).

\subsection{Semantic Quality Rubric}
\label{sec:eval-rubric}

Unlike compliance, settled by a real parser, semantic quality reflects
domain-level judgments a reviewer can inspect rule by rule, so we define it
as an explicit, deterministic, rule-based protocol. Let
$\mathcal{R}=\{R_1,\dots,R_k\}$ be a fixed set of rules, each a function that
takes a STIX bundle and returns a list of violations; a rule $R_i$
\emph{passes} on a bundle $B$ if $R_i(B)=\emptyset$, and the semantic score
for $B$ is the fraction of rules that pass:
\begin{equation}
\mathrm{Semantic}(B) \;=\;
\frac{\bigl|\{R_i\in\mathcal{R} : R_i(B)=\emptyset\}\bigr|}{|\mathcal{R}|}.
\end{equation}
The definition is rule-counting rather than violation-counting: each rule
contributes at most one unit of penalty regardless of how many times it is
violated within the bundle.

Mirage's current rubric uses $|\mathcal{R}|=8$ rules in three classes.
\emph{Type-role rules} require that each STIX object type participate in the
relationships its role implies: a threat-actor must attribute to, use, or target
something (RULE-01)~\cite[\S4.18]{stix21spec}; a malware object must connect to
actor or infrastructure (RULE-02)~\cite[\S4.8]{stix21spec}; an indicator must
indicate a malware, attack-pattern, or actor (RULE-03)~\cite[\S4.7]{stix21spec}; a
campaign must have both an attributed actor and a target
(RULE-04)~\cite[\S4.2]{stix21spec}; and a vulnerability must connect to an
exploitation context (RULE-05)~\cite[\S4.19]{stix21spec}.
\emph{Structural-integrity rules} forbid self-referential edges (RULE-06) and
duplicate triples (RULE-07). The \emph{vocabulary rule} RULE-08 enforces the
whitelist of Section~\ref{sec:method-ir}. The full rule list with sources is
given in the supplementary material.

Three of the eight rules are not informative discriminators among the IR-based
engines: RULE-08 passes by construction for every engine targeting the IR, and
RULE-06 and RULE-07 are degenerate-structure checks those engines satisfy
trivially, so the semantic differences among the IR engines are carried by
RULE-01 through RULE-05. The schema-forced baseline does not pass through the
IR whitelist compiler, so for it RULE-08 is a substantive check rather than a
guarantee. We report the rule-based protocol rather than a human-rater study
because the rules are unambiguous and the score is reproducible from the bundle
alone; an inter-rater study with CTI analysts is a natural extension. A per-rule
decomposition (Table~\ref{tab:perrule-main}, full seven-generator version in
the supplementary material) isolates which rules drive the cross-engine
differences of Section~\ref{sec:results-cross}.

Beyond per-graph scoring, we test whether each generator's output is
\emph{structurally distinguishable} from two reference sets using a
random-forest discriminator on graph-level features. The discriminator is
meaningful only alongside a control, so we include an Erd\H{o}s--R\'enyi (ER)
random-graph control run through the identical pipeline: if it separates random
graphs from a reference as easily as it separates ours, a high AUC against that
reference tells us nothing about realism. As shown below, this control is what
turns one of our results from an apparent success into a correctly identified
negative.

\subsection{Experimental Setup}
\label{sec:eval-setup}
Unless noted otherwise, all engines including the schema-forced baseline use
GPT-4o mini, so comparisons reflect generation method rather than model. We
generate $N=50$ graphs per engine for the cross-engine comparison, reporting
means with $95\%$ bootstrap confidence intervals (10{,}000 resamples) that
quantify sampling variance over the $50$ graphs. The discriminator's two
reference sets are the schema-forced baseline and the MITRE ATT\&CK knowledge
base, and wargaming outcomes are characterized over the same $N=50$ sample.

The discriminator is a random-forest classifier trained on nine graph-level
features: edge density, mean degree, connected-component count, degree
skewness, degree Gini coefficient, triangle, star, and chain motif counts, and
relationship-type entropy, evaluated by $5$-fold stratified cross-validation
(mean AUC against a chance baseline of $0.5$). Because an AUC of $1.0$ can in
principle be reached by overfitting at $N=50$, the cross-validation protocol
and the ER control are jointly the safeguards, and where the AUC saturates we
complement it with the maximum mean discrepancy (MMD) over the same features
against a size-matched ER null drawn from the empirical wargaming size
distribution. The full discriminator protocol, both ER controls, and timing are
detailed in the supplementary material.

\section{Results}
\label{sec:results}

\subsection{Cross-Engine Comparison}
\label{sec:results-cross}

Table~\ref{tab:crossengine} reports the four metrics for every engine. All
engines achieve perfect STIX compliance, confirming the deterministic compiler
delivers valid output regardless of how the graph was generated; the remaining
metrics show a trade-off rather than a single dominant engine. The two causal
engines achieve perfect semantic quality but lower diversity, because an
engagement, or a population evolved from engagements, explores a focused region
of relational behaviour in depth; the Procedural engine inverts this, attaining
high diversity (comparable to the schema-forced baseline) but the lowest
semantic quality, because painting content onto a pre-built skeleton produces
more domain-rule violations than letting structure and meaning co-emerge. This
contrast is evidence for the central design claim: engines in which
relationships exist only because a corresponding event occurred are markedly
more semantically coherent than one that assigns structure first and meaning
afterward. Pairwise Wilcoxon rank-sum tests with Bonferroni correction
($\alpha = 0.05/12$) confirm the picture: Evolutionary and Wargaming are
statistically indistinguishable on all three metrics ($p > 0.05$), while both
are significantly higher than Schema-Forced and Procedural on semantic quality
($p < 0.001$).

\begin{table*}[!t]
\caption{Cross-engine comparison ($N=50$ per engine): mean ($95\%$ bootstrap CI). Superscripts: Evolutionary significantly different from Schema-Forced\,($^S$), Procedural\,($^P$), or Wargaming\,($^W$) (Section~\ref{sec:results-cross}).}
\label{tab:crossengine}
\centering
\setlength{\tabcolsep}{5pt}
\begin{tabular}{lcccc}
\toprule
Engine & Compliance & Structural & Semantic & Diversity ($H$) \\
\midrule
Schema-Forced & 1.00 (1.00\,--\,1.00) & 0.98 (0.98\,--\,0.98) & 0.83 (0.81\,--\,0.86) & 2.73 (2.71\,--\,2.75) \\
Procedural    & 1.00 (1.00\,--\,1.00) & 0.81 (0.77\,--\,0.85) & 0.61 (0.58\,--\,0.64) & 2.66 (2.59\,--\,2.74) \\
Wargaming     & 1.00 (1.00\,--\,1.00) & 0.95 (0.93\,--\,0.97) & 1.00 (1.00\,--\,1.00) & 1.65 (1.61\,--\,1.69) \\
\textbf{Evolutionary} & 1.00 (1.00\,--\,1.00) & 0.98 (0.96\,--\,0.99)$^{S,P}$ & 1.00 (1.00\,--\,1.00)$^{S,P}$ & 1.51 (1.40\,--\,1.61)$^{S,P}$ \\
\midrule
Barab\'asi--Albert & 1.00 (1.00\,--\,1.00) & 0.92 (0.87\,--\,0.96) & 0.72 (0.68\,--\,0.77) & 2.58 (2.29\,--\,2.86) \\
Configuration & 1.00 (1.00\,--\,1.00) & 0.65 (0.59\,--\,0.72) & 0.67 (0.62\,--\,0.72) & 2.55 (2.28\,--\,2.80) \\
Watts--Strogatz & 1.00 (1.00\,--\,1.00) & 0.96 (0.94\,--\,0.99) & 0.72 (0.69\,--\,0.76) & 3.01 (2.82\,--\,3.19) \\
\bottomrule
\end{tabular}
\end{table*}

\begin{table}[!t]
\caption{Per-rule pass rates, total\,/\,vacuous (no node of the rule's subject type), for the five discriminating rules ($N=50$; $48$ for Schema-Forced). Earned: rule-mean of total$-$vacuous over all eight rules; full seven-generator table in the supplementary material.}
\label{tab:perrule-main}
\centering
\small
\setlength{\tabcolsep}{3.5pt}
\begin{tabular}{@{}lccccc@{}}
\toprule
Rule & Sch-F & Proc & War & Evol & Config \\
\midrule
RULE-01 & $.62/.00$ & $.06/.02$ & $.96/.00$ & $1.0/.00$ & $.50/.36$ \\
RULE-02 & $.16/.00$ & $.90/.88$ & $1.0/.04$ & $.98/.24$ & $.38/.30$ \\
RULE-03 & $1.0/1.0$ & $.10/.04$ & $1.0/1.0$ & $1.0/1.0$ & $.44/.38$ \\
RULE-04 & $1.0/1.0$ & $.00/.00$ & $1.0/1.0$ & $1.0/1.0$ & $.38/.38$ \\
RULE-05 & $1.0/1.0$ & $.86/.04$ & $1.0/.04$ & $1.0/.26$ & $.64/.28$ \\
\midrule
Earned & $0.47$ & $0.49$ & $\mathbf{0.73}$ & $\mathbf{0.69}$ & $0.46$ \\
\bottomrule
\end{tabular}
\end{table}

The per-rule decomposition of Table~\ref{tab:perrule-main} settles whether this
gap reflects real type-role coherence or merely the IR engines'
by-construction satisfaction of vocabulary and integrity rules. It is the
former: RULE-08, the whitelist rule, passes on every bundle of every engine,
and restricting attention to the five discriminating type-role rules
(RULE-01 through RULE-05), where no engine has a by-construction advantage,
\emph{widens} the separation: $1.00$ for the causal engines against $0.75$
(Schema-Forced) and $0.38$ (Procedural). The baseline's deficit concentrates
in RULE-01 (a threat-actor with no outgoing relationship, failing on $38\%$
of bundles) and RULE-02 (an isolated malware object, $84\%$), the
ungrounded-relationship failures per-object generation invites and causal
construction precludes. These rates are computed over the released bundle set
($N=48$ for Schema-Forced after two bundles failed validation, $N=50$ for
Procedural) and match the dedicated-regeneration aggregates ($0.84$ and
$0.62$) of the supplement.

The lower block of Table~\ref{tab:crossengine} adds three classic synthetic
graph generators run through the identical compiler and protocol ($N=50$, no
language model; types and admissible verbs assigned at random):
Barab\'asi--Albert, a configuration model matched to the wargaming degree
sequence, and Watts--Strogatz. No public end-to-end synthetic CTI KG generator
exists to compare against (Section~\ref{sec:related}), so these complete the
baseline suite. Their apparent semantic quality ($0.67$ to $0.72$, above
Procedural) is mostly an artifact of \emph{vacuous} passes: a type-role rule
passes trivially on a bundle with no node of its subject type, and excluding
vacuous passes drops earned semantic to $0.46$--$0.50$. RULE-04 is the sharpest
case: every causal engine and the configuration model passes it only vacuously
by emitting no campaigns, while Procedural, the one generator that creates
campaigns, fails it outright. Moreover $41$--$45\%$ of the random generators'
intended edges are rejected by the whitelist at compile time, so the schema
manufactures much of their apparent validity, whereas causal generation never
proposes inadmissible structure at all.

\subsection{Structural Cohesion}
\label{sec:results-struct-decomp}

\begin{figure}[!t]
\centering
\includegraphics[width=0.86\columnwidth]{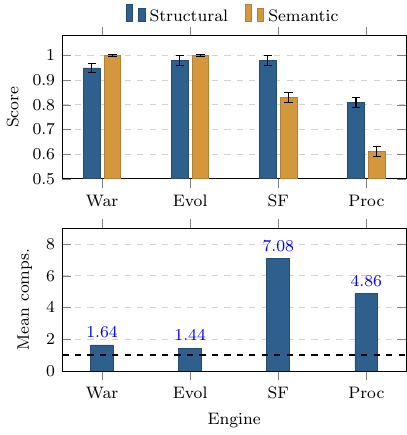}
\caption{Cross-engine results ($N=50$; Table~\ref{tab:crossengine}): headline scores (top) and mean component count (bottom; dashed line marks the single-component ideal).}
\label{fig:cohesion}
\end{figure}

The cross-engine comparison shows Wargaming scoring slightly below
Schema-Forced on structural quality, and Procedural lower still, but
decomposing the score into its multiplicative components,
$\mathrm{Structural} = \mathrm{LCF} \times (1 - \mathrm{Iso})$ (full table in
the supplementary material), shows the headline score is misleading. Isolation
explains essentially none of the gap (near zero across all $100$ causal-engine
bundles); the causal engines produce far fewer disconnected components on
average (Fig.~\ref{fig:cohesion}), and only $40\%$ of wargaming and $20\%$ of
evolutionary bundles lose any structural points at all, against $94\%$ and
$96\%$ for the baselines. The schema-forced score is high only because its many
small disconnected pieces keep the largest-component fraction high: the score
is sensitive to fragment size but blind to fragment count, so we treat
component count as the more informative cohesion measure. On the headline
scores all four systems are close, while the fragment counts separate the
causal engines from both baselines by a factor of three to five.

\subsection{Independent Two-Judge Semantic Evaluation}
\label{sec:results-judge}

To test whether the semantic findings depend on the eight handcrafted
rules, we scored every released bundle with two independent LLM judges from
different providers, \texttt{gpt-4o-2024-08-06} and
\texttt{claude-sonnet-4-5}~\cite{zheng2023judging}: each rates a plain-text
rendering of the graph on plausibility, coherence, and realism ($1$ to $5$,
temperature $0$, three repeats), blind to the generator, the rubric, and the
study design (full tables and figure in the supplementary material). Both judges are highly self-consistent (mean per-bundle repeat
standard deviation $\leq 0.04$), engine differences are significant under
Kruskal--Wallis on every scale ($p < 0.001$), and the judges agree on the
bundle ranking (cross-judge Spearman $\rho = 0.75$ coherence, $0.68$
realism, $0.63$ plausibility).

Three observations. The ranking matches the rubric's, with the causal engines
leading on every scale under both judges and the largest gaps on coherence;
under the stricter Claude judge every schema-forced bundle received the floor
realism score on all three repeats. The judges see what compliance cannot:
schema-forced bundles with invented techniques such as ``gravitational flux
exchange'' score $1.00$ on STIX compliance and the floor on judge realism.
Finally, the judge is itself a metric with a mechanism: its coherence
instruction rewards connectedness, so the classic generators, connected by
construction, score $2.69$ to $2.91$ on GPT-4o coherence despite carrying no
domain logic, and within the two LLM baselines judge scores correlate
\emph{negatively} with the rule-based score (Spearman $\rho = -0.38$,
$p < 0.001$), the vacuous-pass artifact from the other side. No single measure
is trustworthy alone, which is why we triangulate.

\subsection{Finding 1: Structural Distinction from the Schema-Forced Baseline}
\label{sec:results-finding1}
The discriminator separates all three generative engines from the
schema-forced baseline perfectly (AUC $=1.000 \pm 0.000$ under $5$-fold
stratified cross-validation at $N=50$). The point is not that these engines
score higher on quality (Table~\ref{tab:crossengine} shows they do not) but
that they produce a structurally different class of graph. The feature most
driving the separation is the same across all three, average degree
(importance $0.30$ to $0.34$), followed by triangle-motif counts and degree
skewness, consistent with the emission rule in which each resolved compromise
emits a fixed bundle of edges around the affected host. Because the AUC
saturates, we complement it with the MMD over the same nine features with a
permutation test (supplementary material): every pairwise separation is
significant at $p<0.001$, and against the \emph{size-matched} ER control the
wargaming corpus has $\mathrm{MMD}=0.74$ ($p<0.001$), structurally distinct
from a random graph of the same size, not merely from one of a different
size.

\subsection{Finding 2: Separation from MITRE ATT\&CK Is Uninformative as a Realism Signal}
\label{sec:results-finding2}
The discriminator also separates every generative engine from the MITRE
ATT\&CK knowledge base with essentially perfect accuracy
(AUC $=1.000 \pm 0.000$). Taken alone this might look like a realism finding;
the ER control shows it is not, since the same discriminator separates random
ER graphs from ATT\&CK with AUC $=1.000 \pm 0.000$ as well: a manifestly
unrealistic distribution is separated just as perfectly. This instantiates
Corollary~\ref{cor:disc}: the discriminator reads only graph-level features,
the ER control matches them, so by Theorem~\ref{thm:blind} the test cannot
separate coherence from a feature-matched random graph. We do not conclude
ATT\&CK is the unrealistic distribution; with no operational incident graph
in hand the data are equally consistent with our graphs being the outlier.
The recommendation is methodological: AUC against an ATT\&CK-derived reference
is not evidence of operational realism.

\subsection{Finding 3: Downstream Transfer Does Not Favor Structured Graphs, and Why}
\label{sec:results-finding3}
The motivating use of Mirage is as training data for knowledge-graph embedding
(KGE), so the central question is whether a model trained on Mirage transfers
to unseen data. We report a negative result in the sense of
Section~\ref{sec:results-finding2}, evaluated at three levels of
increasing rigour (full table in the supplementary material). \emph{(i)
Transductive, in-distribution}: TransE on the pooled Mirage corpus at the
native UUID level attains $\mathrm{MRR}=0.019\pm0.002$,
$\mathrm{Hits}@1=0.000$; since each bundle mints fresh UUIDs, a transductive
model sees almost no test entity, so this is expected and uninformative.
\emph{(ii) Type-level, cross-corpus}: collapsing every entity to its STIX
type reduces the corpus to nine distinct (type, relation, type) triples and
attains $\mathrm{MRR}=0.335$ but $\mathrm{Hits}@1=0.000$,
$\mathrm{Hits}@10=1.000$ on a MITRE reference that is $86.7\%$ one relation
and $93.6\%$ one tail type, at or below the small-candidate-set random
baseline, hence degenerate. \emph{(iii) Inductive, entity-level} (the sound
test): we trained NodePiece~\cite{nodepiece2022}, which scores entities never
seen in training, on a $40$-bundle training graph and evaluated filtered link
prediction on a $10$-bundle inference graph with disjoint entities but shared
relation vocabulary over three seeds.

Mirage attains $\mathrm{MRR}=0.036\pm0.034$ with $\mathrm{Hits}@1=0.005$
(zero in two of three seeds) against a chance baseline of $0.029$:
indistinguishable from chance, so we do not claim inductive transfer. The
controls are more informative: an ER random-graph corpus transfers at least as
well and on these runs better ($\mathrm{MRR}=0.095$, $\mathrm{Hits}@1=0.029$),
a schema-forced corpus behaves similarly, and the Mirage-to-MITRE stress test
is degenerate as anticipated ($\mathrm{MRR}=0.011$ over $1{,}780$ inference
entities, $92.6\%$ \textsf{uses}). The reason a random corpus is not at a
disadvantage is the point: an inductive KGE model generalizes by learning
relational-neighbourhood regularities that recur across graphs, exactly what an
ER corpus's uniform relation distribution provides, whereas Mirage's
per-scenario specificity does not recur uniformly and so does not help. This is
Corollary~\ref{cor:transfer}: the embedding reads relational statistics, a
random corpus matching them induces the same score law, so none of the three
measurements validates a coherence-by-construction generator. We report the
sound test in full because finding it negative forecloses a measurement the
field would otherwise treat as decisive.

\subsection{Finding 4: Augmentation Gains Are Real but Structure-Agnostic}
\label{sec:results-augmentation}

The preceding finding concerns Mirage-only training; the more common
practical proposal is \emph{augmentation}, adding synthetic graphs to
scarce real training data. We test it directly. From MITRE ATT\&CK
enterprise we extract $20{,}169$ entity-level triples over six object types
and four relation types ($90.3\%$ \textsf{uses}), split $80/10/10$
stratified by relation. Four conditions share identical model, budget, and
splits: real only (A); real plus the $2{,}117$ triples of the $100$
causal-engine bundles (B); real plus a size-matched ER corpus (C); real plus a
size-matched schema-forced corpus (D). Augmentation entities are namespaced
against collision, validation and test contain only real triples, and ranking
is restricted to the $1{,}792$ real entities in every condition. We train
TransE and RotatE (PyKEEN, five seeds), reporting filtered metrics on held-out
real test triples (full protocol in the supplementary material).

The result has two halves. Augmentation helps: under RotatE every source
improves MRR over real-only on every seed ($+8$ to $+12\%$ relative). But the
benefit is not specific to Mirage: the ER control improves as much. Mirage
exceeds the control on MRR on all five seeds, yet the margin is a tenth of the
augmentation effect, reverses on Hits@10, and is absent under TransE, which
gains nothing. The relation-overlap statistics rule out a knowledge-transfer
channel: the fraction of augmentation triples on test-vocabulary relations is
$100\%$ (ER), $44\%$ (Mirage), and $25\%$ (schema-forced), yet the estimates
order the other way, so the gain does not require test-relevant relations,
consistent with a model-dependent regularization effect of added volume. Had we
run only the standard design, B against A, we would have reported a clean
positive at the floor $p$-value; the control unmasks it as the augmentation
instance of Corollary~\ref{cor:transfer}: a reported augmentation benefit is
uninformative about generator quality unless a volume-matched random control
runs alongside, which synthetic-CTI evaluations do not currently include.

\subsection{Robustness: Scaling, Sensitivity, and Model Variation}
\label{sec:results-robustness}
Three robustness studies, reported in full in the supplementary material,
confirm the structural findings are not artifacts of scale, calibration, or
model choice. First, a powered scaling study ($120$ runs) shows wargaming
graph size is set by engagement length, not network size (median $10$ nodes
at $15$ hosts and $9$ at $30$; Mann--Whitney $U=535.5$, $p=0.20$), the
sizing Section~\ref{sec:theory-graph} characterizes analytically; the $N=50$
outcome split is $84\%$ attacker win, $10\%$ eviction, $6\%$ turn-budget
exhaustion. Second, a one-factor-at-a-time sweep over exploit-success
probability, minimum vulnerable-host fraction, and alert threshold ($200$
simulations) holds cohesion stable across the grid, with mean component count
in $[1.50, 2.30]$ everywhere, far below the schema-forced and procedural
baselines, and semantic quality near $1.00$. Third,
rerunning the comparison under GPT-4o (snapshot \texttt{gpt-4o-2024-08-06})
leaves semantic quality and cohesion unchanged (wargaming structural
$\pm0.001$, mean components $1.52$) while diversity moves with the policy
(evolutionary entropy $1.51\to1.64$) and the outcome split is essentially
unchanged ($80/8/12$ vs.\ $84/10/6$), exactly the pattern the construction
predicts since structure is compiler- and referee-authored while variety
follows the policy.

\subsection{Ablation: Which Mechanism Authors the Graph}
\label{sec:results-ablation}

We remove each of the three mechanisms in turn and measure the effect against
the unmodified engine on the metrics of Table~\ref{tab:crossengine}
(Table~\ref{tab:ablation}); each ablation changes exactly one mechanism, so a
change is attributable to the removed component.

\paragraph{Referee (Ablation A)} Resolving every exploit as a success does not
change semantic quality (still $1.00$, since coherence is enforced by the
compiler and whitelist); what collapses is variety. $42$ of $50$ runs produce
the same winning path, relational entropy concentrates at $1.54$ ($95\%$ CI
$[1.53, 1.55]$), entirely below the full-engine baseline ($[1.61, 1.69]$), and
the structural score rises to $1.00$ only because no engagement fragments. The
referee's contribution is the corpus's \emph{variety}, the operational
counterpart of the causal-grounding property of
Section~\ref{sec:theory-referee}.

\paragraph{Evolutionary novelty term (Ablation B)} Selecting by quality alone
does not reduce diversity: per-graph entropy is slightly \emph{higher} without
the novelty term ($1.74$ versus the baseline $1.51$), with semantic and
compliance unchanged. At this budget, rewarding distance from the population
narrows selection rather than widening it, so quality-only pressure explores at
least as broadly; we report this as a measured negative. The quantity the term
optimizes, mean pairwise population distance, was undefined here because each
run retained a single survivor, so we report the per-graph entropy proxy.

\paragraph{Language-model policy (Ablation C)} Replacing the agents with a
uniform random valid-action policy (referee and world model unchanged) leaves
the structural score unchanged within confidence ($0.913$ versus $0.951$,
overlapping intervals): gross connectivity is set by the referee and world
model. Diversity and semantic quality drop measurably: diversity to $1.42$ (CI
$[1.24,1.58]$) against $1.65$ ($[1.61,1.69]$), disjoint, and semantic quality
to $0.98$ ($[0.97,0.99]$) against $1.00$. This supports a division of labour:
the referee and world model author structure, while the policy contributes the
variety and coherence a random policy loses by exhausting the turn budget
without a coherent campaign.

\begin{table}[!t]
\caption{Ablation study: each row removes one mechanism (A, C: wargaming engine, $N=50$; B: evolutionary, $N=25$); metrics as in Table~\ref{tab:crossengine}.}
\label{tab:ablation}
\centering
\small
\setlength{\tabcolsep}{4pt}
\begin{tabular}{@{}lcccc@{}}
\toprule
Configuration & Compl. & Struct. & Sem. & Div.\ ($H$) \\
\midrule
Wargaming (full, $N{=}50$)    & $1.00$ & $0.95$ & $1.00$ & $1.65$ \\
Evolutionary (full, $N{=}50$) & $1.00$ & $0.98$ & $1.00$ & $1.51$ \\
\midrule
A: referee off                & $1.00$ & $1.00$ & $1.00$ & $1.54^{\dagger}$ \\
B: novelty off ($N{=}25$)     & $1.00$ & $0.96$ & $1.00$ & $1.74$ \\
C: random policy ($N{=}50$)   & $1.00$ & $0.91$ & $0.98$ & $1.42$ \\
\bottomrule
\end{tabular}
\end{table}

Taken together, the ablations separate the three mechanisms: structure is
authored by the referee and world model (A, C), the policy contributes the
diversity and coherence a random policy measurably loses (C), and the
novelty term does not widen corpus diversity at this budget (B); the
mechanism the theory identifies as the source of coherence is, by ablation,
also the source of the corpus's structural regularity.

\section{Discussion and Limitations}
\label{sec:discussion}

The results support a bounded claim: Mirage produces STIX-compliant graphs
structurally distinct from schema-forced generation, the causal engines
yield high semantic coherence, and graph size emerges sensibly from the
simulation. We do not claim the graphs are indistinguishable from real
operational intelligence or are embedding-ready; those claims cannot be
evaluated with available references, which is itself one of our findings.

\subsection{The Absence of an Operational Reference}
The scarcity of a realism reference is structural. Operational incident data
records what was \emph{observed} (connections, file writes, indicators),
whereas a coherence-by-construction graph records what was \emph{inferred},
that an actor employs a malware family exploiting a vulnerability class. The
only large public dataset at that abstraction level is ATT\&CK, a curated
knowledge base rather than a corpus of incidents, which Finding~2 shows is
unsuitable as a realism reference; indicator-centric exports and prose
reports do not match a multi-type actor-capability graph in granularity. A
faithful reference therefore requires sanitized incident-response data or a
dedicated extraction pipeline, both left to future work.
Theorem~\ref{thm:blind} states what a valid benchmark must do instead: read
a correlate of coherence that survives feature-matching against a
structure-matched null, for example reconstructing held-out causal
sub-structures within an engagement, the principal direction we identify for
future work. Until it exists, Finding~4 supplies the minimum standard: any
augmentation benefit should carry a volume-matched random-graph control, and
the harness is reference-agnostic, so an operational corpus drops into the
same protocol unmodified.

\subsection{Calibration, Robustness, and the Consistency--Realism Gap}
The eviction outcome is sensitive to the per-event alert weighting, tuned to a
low but nonzero defender-win rate; at $N=50$ each eviction shifts the rate by
two points, and the sweep of Section~\ref{sec:results-robustness} shows the
structural findings are not artifacts of this calibration. Re-running the
comparison under GPT-4o (Section~\ref{sec:results-robustness}) gives the
outcome the construction predicts: compiler- and referee-authored properties
unchanged, diversity shifting with the policy. The world model captures host
services, vulnerabilities, exploitation under uncertainty, lateral movement,
and alert-based detection, but not yet privilege escalation as a distinct
phase, time-correlated detection, or richer credential relationships; because
structure emerges from the world model rather than prompt content, extending it
along these axes yields richer graphs without changing the principle. The
construction guarantees causal consistency \emph{within} the world model; the
gap to real-world realism is the world model's fidelity, since a referee
certifies that every edge records a resolved event, not that the distribution
matches operational intrusions, and no internal argument could
(Remark~\ref{rem:internal-external}). The paradigm is not CTI-specific: any
domain whose artifacts record a process with a definable world model (fraud
chains, intrusion forensics, supply-chain compromise) admits the same
agents-propose, referee-resolves construction, and since
Theorem~\ref{thm:blind} is stated over an abstract coherence property, its
validation pitfalls transfer unchanged.

\subsection{Dual-Use Considerations}
Mirage generates synthetic intelligence graphs, typed records of simulated
engagements, rather than executable exploits, operational tooling, or
target-specific attack plans; we judge the incremental offensive risk modest
relative to public knowledge bases such as ATT\&CK and recommend treating
generated graphs as training artifacts rather than validated intelligence
(expanded in the supplementary material).

\section{Conclusion and Future Work}
\label{sec:conclusion}

We studied generation by simulation as a way to synthesize coherent relational
data: rather than asking a language model to describe a relational artifact, we
simulate a process that produces one and emit the artifact as its causal
record. We instantiated the paradigm for STIX~2.1 CTI knowledge graphs in
Mirage, whose central engine plays a turn-based attacker--defender wargame,
alongside a procedural engine and an evolutionary engine that is, to our
knowledge, the first to apply language-model genetic operators to structured
graph genomes; one deterministic compiler makes dangling and malformed output
unrepresentable, with proved referential closure, causal grounding, and
identifier stability, and an ablation localizing structure to the referee and
world model and variety to the policy.

We prove that a validator reading any feature set a structure-matched random
graph can reproduce is uninformative about coherence for a
coherence-by-construction generator (Theorem~\ref{thm:blind}), and the
empirical evaluation realizes the theorem four times over: separation from
MITRE ATT\&CK, type-level transfer, inductive entity-level transfer, and
augmentation of real training data each prove uninformative, with an identified
mechanism in each case, while two independent LLM judges reproduce the rubric's
engine ranking. We therefore claim not demonstrated embedding-readiness but a
map of which downstream validations are uninformative for such generators and
why, a paradigm with formal guarantees any such generator can adopt, and the
benchmark still needed: a transfer task rewarding per-scenario coherence rather
than neighbourhood regularity, paired with the sanitized operational reference
the CTI setting lacks.

{\footnotesize
\bibliographystyle{IEEEtran}
\bibliography{mirage}
}

\clearpage
\setcounter{section}{0}
\setcounter{subsection}{0}
\setcounter{table}{0}
\setcounter{figure}{0}
\renewcommand{\thesection}{S-\Roman{section}}
\renewcommand{\thesubsection}{\Alph{subsection}}
\renewcommand{\thetable}{S-\Roman{table}}
\renewcommand{\thefigure}{S-\arabic{figure}}

\begin{center}
{\large\bfseries Supplementary Material}\\[2pt]
{\normalsize Coherence by Construction: Synthesizing CTI Knowledge Graphs\\
and the Formal Limits of Coherence Validation}
\end{center}

\vspace{2mm}
\noindent
This supplement collects material referenced from the main paper: the extended theoretical remarks and the per-engine complexity derivations (Section~\ref{ssec:proofs}); the configuration and reference tables for the discriminator, the knowledge-graph-embedding experiment, and the semantic rubric, together with the per-engine tables relocated from the main paper for length (Section~\ref{ssec:tables}); the wargaming single-turn pseudocode, calibrated parameters, and engagement-size and parameter-sweep plots (Section~\ref{ssec:wargaming-supp}); the full inductive-transfer and augmentation results with the relation-overlap statistics (Section~\ref{ssec:downstream-supp}); the two-judge evaluation protocol, verbatim judge prompt, and full score tables (Section~\ref{ssec:judge-supp}); the per-rule decomposition with vacuous passes for all seven generators (Section~\ref{ssec:vacuous-supp}); and the verbatim language-model prompts (Section~\ref{ssec:prompts}). Section and table numbers prefixed with ``S'' are internal to this supplement; unprefixed section references point to the main paper.

\vspace{1em}

\section{Extended Theoretical Remarks and Derivations}
\label{ssec:proofs}

The propositions, the central theorem, and the principal proofs appear in the main paper (Section~V). This section gives the full proofs of the two compiler-correctness propositions (referential closure and identifier stability), whose statements are in the main paper, together with the supporting remarks and derivations referenced from there.

\paragraph{Proof of referential closure (main paper Proposition~1)}
Let $r$ be emitted from relationship statement $t \in T$. Emission of $r$ occurs only in the emission phase and only after the minting phase has completed. The minting phase iterates over $T$ and, for each $t$, inserts both $(\sigma_s,\iota_s)$ and $(\sigma_t,\iota_t)$ into the domain of $\mu$ if not already present, so at the end of the minting phase $\{(\sigma_s,\iota_s),(\sigma_t,\iota_t)\} \subseteq \mathrm{dom}(\mu)$ for every $t \in T$, including the $t$ that produced $r$. The emission phase creates one STIX Domain Object for every key in $\mathrm{dom}(\mu)$, so in particular it creates objects with UUIDs $\mu(\sigma_s,\iota_s)$ and $\mu(\sigma_t,\iota_t)$, and the references of $r$ are constructed from exactly these two UUIDs. Because $r$ was arbitrary, the claim holds for every emitted relationship. The whitelist filter can only remove relationships from emission and never add one, so it does not affect the conclusion.

\paragraph{Proof of identifier stability (main paper Proposition~2)}
The map $\mu$ assigns a key a UUID only on its first occurrence, under a guard that tests membership in $\mathrm{dom}(\mu)$; subsequent occurrences leave $\mu$ unchanged, so $\mu$ is a well-defined function and every use of a node key in the emission phase reads the same value, giving intra-compilation stability. For the inter-compilation claim, each fresh UUID draws $122$ random bits (version~4), so by a union bound the probability that any of the $|\mathcal{N}_1| \cdot |\mathcal{N}_2|$ cross-pairs between two compilations collides is at most $|\mathcal{N}_1|\,|\mathcal{N}_2|\,2^{-122}$, negligible for any corpus of practical size.

\paragraph{Referee invariants verified by inspection}
The referee maintains the two invariants used implicitly above: a host's compromised flag is set only upon a successful exploit and never cleared except by isolation, and a host's alert counter is monotonically nondecreasing within an engagement. Both follow by inspection of the referee implementation, since the only writes to these fields are the increment on a detected event and the set-on-success of the compromise flag.

\paragraph{Per-engine complexity derivations}
The closed-form statements quoted in the main paper (Section~V-C) follow from the per-turn and per-generation operation counts. For the wargaming engine, let $H$ be the number of hosts and $B=\max(25,\lfloor 1.5H\rfloor)$ the turn budget (main paper, Section~IV-D). Each turn issues a constant number of language-model calls and performs referee work bounded by the per-turn neighborhood expansion, which is $O(H)$ in the worst case. The engine therefore runs in $O(B\cdot H) = O(H^2)$ worst-case referee time exclusive of language-model latency, and emits $O(B)$ relationship statements, since each turn contributes a bounded edge bundle. Procedural generation (Engine~A) is dominated by the scale-free construction at $O(n\log n)$ for $n$ nodes plus one batched language-model call. The evolutionary engine (Engine~C) over $G$ generations with population size $\pi$ performs $O(G\pi)$ language-model operator calls and, for the global novelty computation, $O(\pi^2)$ pairwise distance evaluations per generation, each linear in graph size, for a total structural cost $O(G\pi^2\bar{s})$ at mean graph size $\bar{s}$, exclusive of model latency.

\paragraph{Reach of the characterization}
The two structural properties stated in the main paper (Section~V-D) characterize the \emph{generative} distribution, the structure the process must produce given its rules. They do not establish that this distribution matches that of real operational CTI, and the causal-grounding proposition, together with the internal-versus-external-realism remark in the main paper, makes clear why no purely internal argument could: causal consistency within the world model is independent of the world model's fidelity to reality. A full analytic parameter-to-structure map, giving expected component counts and motif frequencies as closed-form functions of the exploit-success probabilities, the alert threshold, and the vulnerability fraction, would extend this characterization, and we identify it as the natural completion of this analysis. The empirical sensitivity sweep it complements is reported in the main paper (Section~VIII-H) and tabulated in Table~\ref{stab:sweep}. For Engine~A, we additionally note that type assignment is a function of in- and out-degree and therefore does not alter the degree sequence of the underlying construction.

\paragraph{On the evolutionary loop}
For Engine~C the guard property quoted in the main paper holds by the same argument as the referee invariants above: selection is deterministic and admits a candidate only if it clears the structural sanity floor, the floor is checked before admission, so the population at every generation consists of structurally valid graphs and the loop cannot drift into malformed regions of the search space regardless of what the language-model operator proposes. This is a safety invariant, not a convergence guarantee. We do not claim a diversity-preservation or mode-collapse-avoidance theorem. Such a claim would require a formal model of the operator's proposal distribution, which is a language model and is not analytically characterized, and it would have to be matched by the corpus-level diversity measurement the main paper identifies as unfinished. We therefore present the structural validity invariant as the formal content of Engine~C and treat convergence and the preservation of diversity as open questions.

\section{Configuration and Reference Tables}
\label{ssec:tables}

This section collects the configuration tables for the discriminator
(Table~\ref{stab:disc}) and the knowledge-graph-embedding experiment
(Table~\ref{stab:kge}), the full rule list for the semantic rubric
(Table~\ref{stab:rubric}), and the per-engine tables relocated from the
main paper for length: the full eight-rule per-engine pass-rate table
(Table~\ref{stab:perrule}), the structural-quality decomposition
(Table~\ref{stab:decomp}), the parameter sensitivity sweep
(Table~\ref{stab:sweep}), the cross-model comparison
(Table~\ref{stab:crossmodel}), the maximum-mean-discrepancy table
(Table~\ref{stab:mmd}), and the positioning comparison of the closest
related systems (Table~\ref{stab:positioning}). These are reference tables for the
protocols and results described in the main paper (Sections~IV, VII,
and~VIII).

\begin{table}[!t]
\caption{Random-forest discriminator: configuration and graph-level
features. Referenced from the main paper, Section~VII-C.}
\label{stab:disc}
\centering
\small
\setlength{\tabcolsep}{5pt}
\begin{tabular}{@{}ll@{}}
\toprule
Property & Value \\
\midrule
Classifier & \texttt{RandomForestClassifier} (scikit-learn) \\
Number of trees & 100 \\
Maximum depth & unbounded \\
Split criterion & Gini impurity \\
Evaluation protocol & 5-fold stratified cross-validation \\
Reported statistic & Mean AUC $\pm$ std across folds \\
Chance baseline & $\mathrm{AUC}=0.5$ \\
\bottomrule
\multicolumn{2}{@{}p{0.95\columnwidth}@{}}{\footnotesize Graph-level
features (9): edge density, mean degree, number of connected components,
degree skewness, degree Gini coefficient, triangle motif count, star motif
count, chain motif count, and Shannon entropy of the relationship-type
distribution.}
\end{tabular}
\end{table}

\begin{table}[!t]
\caption{Knowledge-graph-embedding \emph{transfer} experiment (Finding~3) configuration; the augmentation experiment of Finding~4 uses $300$ epochs and five seeds, and is configured separately (Section~\ref{ssec:downstream-supp}). The type-level training corpus is the entity-level corpus with every entity collapsed to its STIX object type, which reduces it to nine distinct type-level triples; this small type vocabulary is central to the interpretation in the main paper (Section~VIII-F).}
\label{stab:kge}
\centering
\small
\setlength{\tabcolsep}{4pt}
\begin{tabular}{@{}lp{4.6cm}@{}}
\toprule
Property & Value \\
\midrule
Library & PyKEEN (TransE and NodePiece) \\
Embedding dimension & 128 \\
Training epochs & 100 \\
Batch size & 256 \\
Optimizer & Adam, learning rate $10^{-3}$ \\
Evaluation & Filtered, mean over 3 seeds \\
Transductive split & 80/10/10 train/val/test (in-distribution) \\
Inductive split & 40-bundle train graph / 10-bundle inference graph \\
Entity-level corpus & 996 triples, 824 entities, 5 relations \\
Type-level corpus & 9 unique (type, relation, type) triples \\
MITRE coverage (type-level) & 11{,}790/21{,}025 triples (56.1\%) \\
\bottomrule
\end{tabular}
\end{table}

\begin{table}[!t]
\caption{Semantic-quality rubric: rules and their sources. Each rule is a
deterministic check over a STIX bundle and contributes one unit to the score
denominator $|\mathcal{R}|=8$. Referenced from the main paper,
Section~VII-B.}
\label{stab:rubric}
\centering
\small
\setlength{\tabcolsep}{4pt}
\begin{tabular}{@{}lp{4.7cm}l@{}}
\toprule
ID & What it checks & Source \\
\midrule
RULE-01 & threat-actor has $\geq 1$ attributed-to/uses/targets edge & \S4.18 \\
RULE-02 & malware connects to actor/tool/infrastructure & \S4.8 \\
RULE-03 & indicator indicates malware/attack-pattern/actor & \S4.7 \\
RULE-04 & campaign has attributed actor and target & \S4.2 \\
RULE-05 & vulnerability has exploits/related edge & \S4.19 \\
RULE-06 & no self-referential relationships & integrity \\
RULE-07 & no duplicate (src, type, tgt) triples & integrity \\
RULE-08 & relationship verb in (src, tgt)-type whitelist & rel.\ vocab. \\
\bottomrule
\multicolumn{3}{@{}p{0.95\columnwidth}@{}}{\footnotesize Sources are clauses
of the STIX 2.1 specification unless marked ``integrity.''}
\end{tabular}
\end{table}

\begin{table}[!t]
\caption{Per-rule semantic pass rates by engine (relocated from the main
paper). Each cell is the fraction of bundles on which that rule passes. The
causal engines (Wargaming, Evolutionary) pass all eight rules on every
bundle, entailed by their uniform semantic score of $1.00$. The
Schema-Forced and Procedural figures are from a regeneration under the
identical engine and model; their All-8 aggregates ($0.84$ and $0.62$)
match the cross-engine scores of $0.83$ and $0.61$ in the main paper to
within the regeneration sample (Schema-Forced $N=48$ after two bundles
failed schema validation at generation, Procedural $N=50$). Starred rules
(RULE-01 through RULE-05) are the discriminating type-role rules; RULE-06
through RULE-08 pass by construction for the IR engines. ``Disc.-5''
averages the starred rules only.}
\label{stab:perrule}
\centering
\small
\setlength{\tabcolsep}{3pt}
\begin{tabular}{@{}lcccc@{}}
\toprule
Rule & Schema-Forced & Procedural & Wargaming & Evolutionary \\
\midrule
RULE-01$^{*}$ & $0.60$ & $0.06$ & $1.00$ & $1.00$ \\
RULE-02$^{*}$ & $0.13$ & $0.90$ & $1.00$ & $1.00$ \\
RULE-03$^{*}$ & $1.00$ & $0.10$ & $1.00$ & $1.00$ \\
RULE-04$^{*}$ & $1.00$ & $0.00$ & $1.00$ & $1.00$ \\
RULE-05$^{*}$ & $1.00$ & $0.86$ & $1.00$ & $1.00$ \\
RULE-06 & $1.00$ & $1.00$ & $1.00$ & $1.00$ \\
RULE-07 & $1.00$ & $1.00$ & $1.00$ & $1.00$ \\
RULE-08 & $1.00$ & $1.00$ & $1.00$ & $1.00$ \\
\midrule
All-8 & $0.84$ & $0.62$ & $1.00$ & $1.00$ \\
Disc.-5$^{*}$ & $0.75$ & $0.38$ & $1.00$ & $1.00$ \\
\bottomrule
\end{tabular}
\end{table}

\begin{table}[!t]
\caption{Decomposition of the structural-quality score per engine (relocated from the main paper). The score factors as $\mathrm{Structural} = \mathrm{LCF} \times (1-\mathrm{Iso})$, where LCF is the largest-connected-component fraction, and Iso is the fraction of degree-$0$ nodes. Wargaming and Evolutionary were re-analyzed from cached bundles. Per-bundle isolation rates were not cached for Schema-Forced and Procedural, so Iso is reported as N/A, and the score equals the LCF as a lower bound. ``Mean comps.''\ is the mean number of connected components, the cohesion indicator used in the main paper.}
\label{stab:decomp}
\centering
\small
\setlength{\tabcolsep}{5pt}
\begin{tabular}{@{}lcccc@{}}
\toprule
Engine & Struct. & LCF & Iso.\ rate & Mean comps. \\
\midrule
Schema-Forced & $0.98$ & $0.98$ & N/A & $7.08$ \\
Procedural & $0.81$ & $0.81$ & N/A & $4.86$ \\
Wargaming & $0.95$ & $0.95$ & $0.000$ & $1.64$ \\
Evolutionary & $0.97$ & $0.97$ & $0.001$ & $1.44$ \\
\bottomrule
\end{tabular}
\end{table}

\begin{table}[!t]
\caption{Parameter sensitivity sweep ($N=20$ per cell, $15$ hosts,
reachability-aware selection, GPT-4o mini), relocated from the main paper.
One factor is varied per block, while the others are held at the baseline
(starred). ``Comp.''\ is the mean number of connected components, the
cohesion indicator (lower is more cohesive); the Schema-Forced and
Procedural baselines sit at $7.08$ and $4.86$ respectively
(Table~\ref{stab:decomp}). ``Win\%'' is the attacker-win fraction;
``Size'' is the median node count; ``Div.''\ is Shannon entropy $H$.}
\label{stab:sweep}
\centering
\small
\setlength{\tabcolsep}{4pt}
\begin{tabular}{@{}lcccccc@{}}
\toprule
Value & Win\% & Size & Comp. & Struct. & Sem. & Div. \\
\midrule
\multicolumn{7}{@{}l}{\emph{Exploit-success probability}} \\
$0.50$ & $35$ & $9$ & $1.65$ & $0.91$ & $0.99$ & $1.67$ \\
$0.70$ & $65$ & $10$ & $1.80$ & $0.89$ & $0.99$ & $1.70$ \\
$0.85^{*}$ & $75$ & $10$ & $1.50$ & $0.94$ & $1.00$ & $1.68$ \\
$0.95$ & $45$ & $14$ & $2.30$ & $0.80$ & $0.96$ & $1.24$ \\
\midrule
\multicolumn{7}{@{}l}{\emph{Minimum vulnerable-host fraction}} \\
$0.30$ & $65$ & $12$ & $2.00$ & $0.87$ & $0.98$ & $1.41$ \\
$0.50$ & $80$ & $10$ & $1.75$ & $0.94$ & $1.00$ & $1.68$ \\
$0.70^{*}$ & $75$ & $10$ & $1.50$ & $0.94$ & $1.00$ & $1.68$ \\
$0.90$ & $70$ & $11$ & $1.65$ & $0.89$ & $0.99$ & $1.54$ \\
\midrule
\multicolumn{7}{@{}l}{\emph{Alert threshold}} \\
$2$ & $55$ & $10$ & $1.65$ & $0.89$ & $0.98$ & $1.57$ \\
$3^{*}$ & $75$ & $10$ & $1.50$ & $0.94$ & $1.00$ & $1.68$ \\
$4$ & $75$ & $13$ & $2.05$ & $0.89$ & $0.99$ & $1.61$ \\
$5$ & $85$ & $13$ & $1.75$ & $0.92$ & $1.00$ & $1.66$ \\
\bottomrule
\multicolumn{7}{@{}p{0.95\columnwidth}@{}}{\footnotesize The mean
connected-component count stays within $[1.50, 2.30]$ at every setting, far
below the two baselines, and semantic quality stays at or near $1.00$
throughout. The interpretation of the three high-component cells is given
in the main paper (Section~VIII-H).}
\end{tabular}
\end{table}

\begin{table}[!t]
\caption{Cross-model comparison ($N=50$ per engine), relocated from the
main paper. The policy model is changed from GPT-4o mini to GPT-4o
(\texttt{gpt-4o-2024-08-06}) with all else fixed; the API response model
field confirmed the swap. Mini values are the main paper cross-engine
values; GPT-4o values carry $95\%$ bootstrap CIs (1000 resamples).
``Comp.''\ is the mean number of connected components under GPT-4o, shown
on the diversity row. The Wargaming and Procedural rows are clean
single-variable swaps; the Evolutionary genetic operators were already on
GPT-4o in the baseline ($\dagger$), so only its wargaming-seed call
differs.}
\label{stab:crossmodel}
\centering
\small
\setlength{\tabcolsep}{4pt}
\begin{tabular}{@{}llccc@{}}
\toprule
Engine & Metric & mini & GPT-4o ($95\%$ CI) & Comp. \\
\midrule
Procedural & Compl. & $1.00$ & $1.00$ & \\
 & Struct. & $0.81$ & $0.83$ $[0.79, 0.86]$ & \\
 & Sem. & $0.61$ & $0.61$ $[0.59, 0.63]$ & \\
 & Div. & $2.66$ & $2.59$ $[2.51, 2.67]$ & $4.90$ \\
\midrule
Wargaming & Compl. & $1.00$ & $1.00$ & \\
 & Struct. & $0.95$ & $0.95$ $[0.93, 0.97]$ & \\
 & Sem. & $1.00$ & $1.00$ & \\
 & Div. & $1.65$ & $1.67$ $[1.62, 1.71]$ & $1.52$ \\
\midrule
Evol.$^{\dagger}$ & Compl. & $1.00$ & $1.00$ & \\
 & Struct. & $0.98$ & $0.96$ $[0.94, 0.98]$ & \\
 & Sem. & $1.00$ & $1.00$ & \\
 & Div. & $1.51$ & $1.64$ $[1.57, 1.69]$ & $1.40$ \\
\bottomrule
\end{tabular}
\end{table}

\begin{table}[!t]
\caption{Maximum mean discrepancy (MMD) between corpora over the nine
graph-level features, relocated from the main paper (Finding 1,
Section~VIII). An RBF kernel with the median-bandwidth heuristic and a
permutation test ($B=1000$) are used. Features are $z$-scored on each
comparison's pooled sample, so the bandwidth is set per comparison and MMD
values are calibrated \emph{within} a comparison (the permutation
$p$-value) but only suggestive when ranked \emph{across} comparisons. ``ER
(matched)'' draws node counts from the empirical wargaming size
distribution; this is the size-matched control referenced in the main
paper, distinct from the size-unmatched control used for the AUC results. A
self-comparison (one corpus split in half) gives $\mathrm{MMD}^2 = -0.015$
($p=0.66$), confirming the estimator returns zero for indistinguishable
samples.}
\label{stab:mmd}
\centering
\small
\setlength{\tabcolsep}{4pt}
\begin{tabular}{@{}lccc@{}}
\toprule
Comparison & MMD & $p$ & $(n_X, n_Y)$ \\
\midrule
Wargaming vs.\ Schema-Forced & $0.96$ & $<0.001$ & $(50, 50)$ \\
Wargaming vs.\ MITRE ATT\&CK & $0.81$ & $<0.001$ & $(50, 164)$ \\
ER vs.\ MITRE ATT\&CK & $0.85$ & $<0.001$ & $(164, 164)$ \\
Wargaming vs.\ ER (matched) & $0.74$ & $<0.001$ & $(50, 50)$ \\
Procedural vs.\ ER (matched) & $0.83$ & $<0.001$ & $(50, 50)$ \\
Evolutionary vs.\ ER (matched) & $0.66$ & $<0.001$ & $(50, 50)$ \\
Schema-Forced vs.\ ER (matched) & $0.77$ & $<0.001$ & $(50, 50)$ \\
\bottomrule
\end{tabular}
\end{table}

\begin{table}[!t]
\caption{Position of Mirage among the closest related systems, relocated from the main paper (Section~II). ``Structure authority'' is who decides which relationships enter the produced artifact. In cyber-range systems, the agents themselves emit the actions that constitute the trace, whereas in Mirage, agents propose intent, and a deterministic referee resolves outcomes against an explicit world model. The smart-grid row records the finding that unconstrained language-model action selection is unreliable for this
purpose.}
\label{stab:positioning}
\centering
\small
\setlength{\tabcolsep}{3pt}
\begin{tabular}{@{}>{\raggedright\arraybackslash}p{1.9cm}>{\raggedright\arraybackslash}p{1.45cm}>{\raggedright\arraybackslash}p{1.7cm}>{\raggedright\arraybackslash}p{1.35cm}>{\raggedright\arraybackslash}p{1.5cm}@{}}
\toprule
System & Purpose & Output artifact & Agent type & Structure authority \\
\midrule
CyberBattle\-Sim & Train RL agents & Trained policy + metrics & RL & Agent \\
CybORG++ & Train RL agents & Trained policy + metrics & RL & Agent \\
CyGIL & Sim/ emulation transfer & Trained policy + simulator & RL & Agent \\
Agrawal et al. & Train RL/ML agents & Trained policy + metrics & RL/ML & Agent \\
Sen et al. & Attack-stage generation & Action sequences & LLM (unconstrained) & Agent (unreliable) \\
Mirage (this work) & Data generation & STIX 2.1 knowledge graph & LLM & Determin\-istic referee \\
\bottomrule
\end{tabular}
\end{table}

\section{Wargaming Procedure, Parameters, and Engagement Distributions}
\label{ssec:wargaming-supp}

\begin{figure*}[!t]
\centering
\begin{tikzpicture}
\begin{groupplot}[
    group style={group size=3 by 1, horizontal sep=28pt,
                 ylabels at=edge left, yticklabels at=edge left},
    width=0.30\textwidth, height=0.22\textwidth,
    ymin=0, ymax=5.6,
    ylabel={Mean components},
    tick label style={font=\footnotesize},
    label style={font=\footnotesize},
    title style={font=\footnotesize},
    ymajorgrids, major grid style={dashed, gray!25},
]
\nextgroupplot[title={Exploit succ.\ prob.}, xtick={0.5,0.7,0.9}]
\addplot[mred, dashed, thick, domain=0.46:0.99] {4.86};
\addplot+[mblue, mark=*, mark size=2pt, thick, mark options={fill=mblue}]
    coordinates {(0.50,1.65) (0.70,1.80) (0.85,1.50) (0.95,2.30)};
\nextgroupplot[title={Min.\ vuln.\ fraction}, xtick={0.3,0.5,0.7,0.9}]
\addplot[mred, dashed, thick, domain=0.26:0.94] {4.86};
\addplot+[mblue, mark=*, mark size=2pt, thick, mark options={fill=mblue}]
    coordinates {(0.30,2.00) (0.50,1.75) (0.70,1.50) (0.90,1.65)};
\nextgroupplot[title={Alert threshold}, xtick={2,3,4,5}]
\addplot[mred, dashed, thick, domain=1.8:5.2] {4.86};
\addplot+[mblue, mark=*, mark size=2pt, thick, mark options={fill=mblue}]
    coordinates {(2,1.65) (3,1.50) (4,2.05) (5,1.75)};
\end{groupplot}
\end{tikzpicture}
\caption{Parameter sensitivity of structural cohesion ($N=20$ per cell,
$200$ simulations): mean component count under one-factor-at-a-time
variation. Dashed line: Procedural baseline ($4.86$; Schema-Forced is
$7.08$, off scale). The count stays within $[1.50, 2.30]$ at every setting.}
\label{fig:sweep}
\end{figure*}
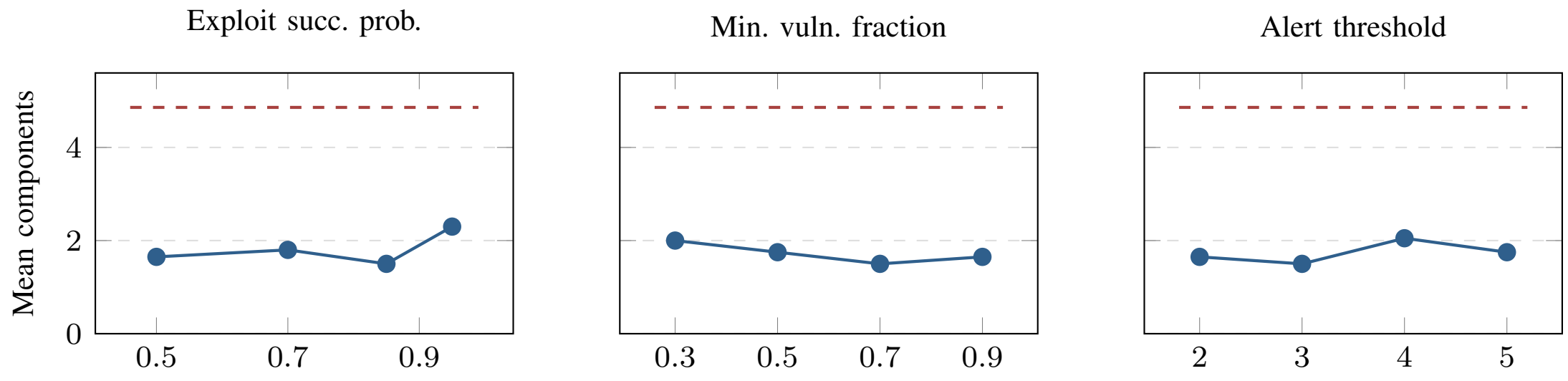

The single-turn procedure of the wargaming engine is presented in Algorithm~1 in the main paper (Section~IV-D); Table~\ref{tab:params} here lists the calibrated simulation parameters it uses. Fig.~\ref{fig:scaling} plots the per-cell engagement-size distributions of the powered scaling study (main paper, Section~VIII-H), and Fig.~\ref{fig:sweep} plots the one-factor-at-a-time parameter sweep whose full numbers are in Table~S-VI.

\subsection{Detection and Eviction}
The defender does not see the attacker directly; it accumulates alerts. Three
kinds of event each raise the alert level of the host involved, a failed
exploit attempt, a successful compromise, and a scan from an already-compromised
foothold, and when a host's alerts reach a fixed threshold the defender may
isolate it, removing it from the attacker's reach. This mirrors how a monitored
network surfaces an intrusion: an attacker generates signal whenever it probes,
breaks in, or moves laterally. The eviction outcome is sensitive to how these
events are weighted, so we treat the alert weighting as a calibrated parameter
(main paper, Section~IX).

\subsection{Termination and Emergent Size}
An engagement ends in one of three ways: the attacker compromises a designated
crown-jewel host, the defender evicts the attacker from every foothold, or a
turn budget of $\max(25, \lfloor 1.5\,n \rfloor)$ turns for $n$ hosts is
exhausted. Two choices keep terminations meaningful: the crown jewel is chosen
from hosts reachable from the entry point along directed edges rather than by
raw degree (in a directed scale-free graph the highest-degree nodes are
typically unreachable sinks), and the entry host must have enough outbound
connectivity for lateral movement. Graph size and shape then \emph{emerge} from
how the engagement plays out rather than from a parameter. These are modeling
choices tuned to a design target of a low-but-nonzero defender-win rate, so the
outcome split of the main paper (Section~VIII-H) is a target achieved rather
than a discovered behaviour; with \texttt{MIN\_VULN\_FRACTION}~$=0.70$ and a
$0.85$ exploit-success probability this is a deliberately attacker-favourable,
soft-target environment, and the sweep shows component cohesion holds across it.


\begin{table}[!t]
\caption{Wargaming simulation parameters (calibrated), relocated from the
main paper for length.}
\label{tab:params}
\centering
\small
\setlength{\tabcolsep}{4pt}
\begin{tabular}{@{}llp{2.4cm}@{}}
\toprule
Parameter & Value & Meaning \\
\midrule
\texttt{\footnotesize VULNERABLE\_HOST\_SUCCESS\_PROB} & $0.85$ & P(exploit $\mid$ vulnerable) \\
\texttt{\footnotesize PATCHED\_HOST\_SUCCESS\_PROB}    & $0.05$ & P(exploit $\mid$ patched) \\
\texttt{\footnotesize ALERT\_THRESHOLD}               & $3$    & alerts before isolation \\
\texttt{\footnotesize MIN\_VULN\_FRACTION}            & $0.70$ & min.\ hosts with $\geq 1$ CVE \\
\texttt{\footnotesize TURNS\_PER\_HOST}               & $1.5$  & turn budget $\max(25, \lfloor 1.5\,n \rfloor)$ for $n$ hosts \\
Entry out-degree               & $\geq 2$ & min.\ out-edges from entry \\
Failed/success/scan alert      & $+1$ ea. & three-event detection \\
\bottomrule
\end{tabular}
\end{table}

\begin{figure*}[!t]
\centering
\begin{tikzpicture}
\begin{axis}[
    width=0.72\textwidth,
    height=0.26\textwidth,
    boxplot/draw direction=y,
    ylabel={Graph size (nodes)},
    xlabel={Selection regime / network size},
    ymin=0, ymax=72,
    xtick={1,2,3,4},
    xticklabels={{R/15},{R/30},{D/15},{D/30}},
    x tick label style={font=\small},
    tick label style={font=\small},
    label style={font=\small},
    ymajorgrids, major grid style={dashed, gray!30},
]
\addplot+[boxplot prepared={lower whisker=3, lower quartile=9, median=10,
    upper quartile=21.8, upper whisker=35}, mblue, fill=mblue!18,
    thick, mark options={mblue}] coordinates {};
\addplot+[boxplot prepared={lower whisker=3, lower quartile=6, median=9,
    upper quartile=13, upper whisker=23.5}, mblue, fill=mblue!18,
    thick, mark options={mblue}] coordinates {(0,70) (0,61)};
\addplot+[boxplot prepared={lower whisker=3, lower quartile=9, median=12,
    upper quartile=14.8, upper whisker=23.4}, mblue!70!black, fill=mblue!8,
    thick, dashed, mark options={mblue!70!black, solid}] coordinates {(0,29)};
\addplot+[boxplot prepared={lower whisker=3, lower quartile=9, median=13,
    upper quartile=21.8, upper whisker=40.9}, mblue!70!black, fill=mblue!8,
    thick, dashed, mark options={mblue!70!black, solid}] coordinates {(0,46) (0,54) (0,60)};
\end{axis}
\end{tikzpicture}
\caption{Wargaming graph size by condition ($n=30$ per cell):
\textsf{R}=reachability-aware, \textsf{D}=naive degree-based selection, at
15 and 30 hosts. Boxes: IQR and median; whiskers: $1.5\times$IQR; dots:
outliers. The right skew is the engagement-driven sizing characterized
analytically in the main paper (Section~V-D).}
\label{fig:scaling}
\end{figure*}
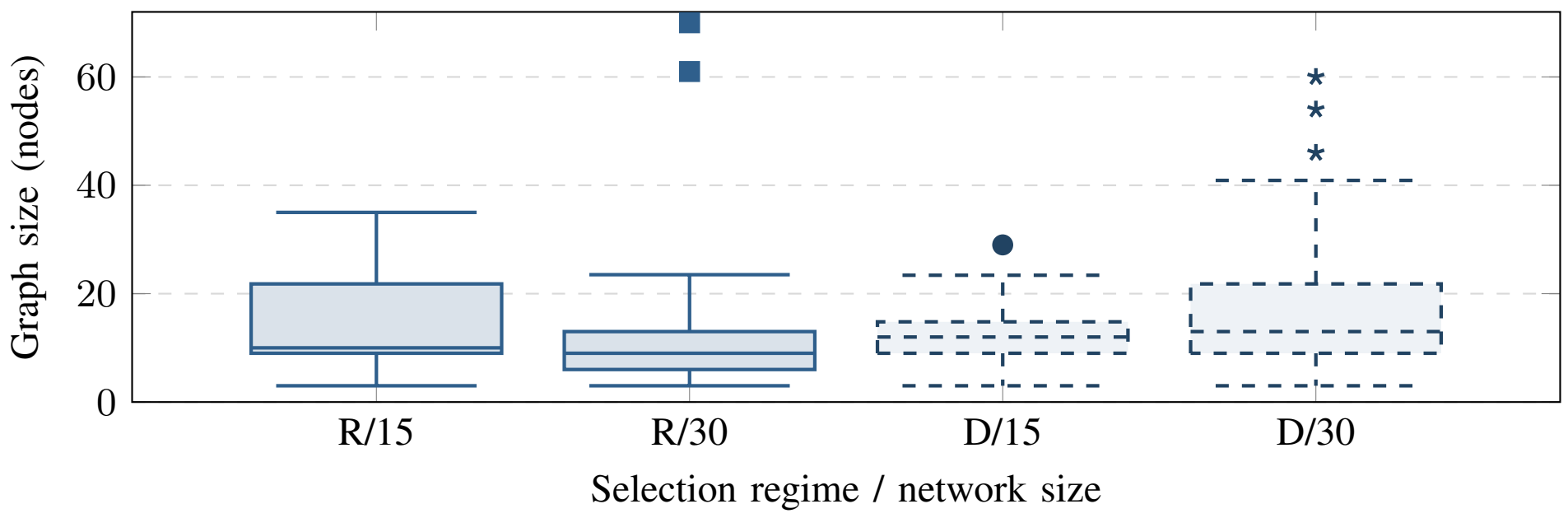

\section{Extended Related Work: Downstream Consumers and Evolutionary Lineage}
\label{ssec:relwork-supp}

This section expands the related-work discussion of the main paper
(Section~II) on the two strands most peripheral to the core contribution: the
downstream KG-embedding consumers that motivate the work, and the lineage of
language-model-driven evolutionary operators.

A growing literature pits language models against each other in red- and
blue-team roles, but most of it targets the safety of the language model
itself: Ma et al.\ study multi-round red-team games, identify mode collapse as
the central obstacle, and respond with a diverse attacker population under
explicit diversity pressure~\cite{evolvingredteam2024}, and Abuadbba et al.\
survey language models across red- and blue-team functions and document where
they remain brittle~\cite{rethinkingredblue2025}. In Mirage, by contrast, the
agents act against a simulated network and the model is the policy rather than
the victim; the mode-collapse finding nonetheless applies and motivates the
evolutionary engine (main paper, Section~IV-E), where corpus-level diversity is
a design goal rather than a demonstrated outcome.

The downstream consumers of CTI knowledge graphs supply this work's motivation
and report being short of data: Fieblinger et al.\ apply link prediction but
report scaling as open~\cite{fieblinger2024}, Zhang et al.\ develop edge
propagation where training data is limited~\cite{zhang2023edge}, Bratsas et
al.\ catalogue the line in a systematic review~\cite{bratsas2024kgsurvey}, and
LineaRE is a representative embedding model these consumers rely
on~\cite{peng2020lineare}. The evolutionary engine has a separate lineage:
using a language model as a genetic operator with deterministic selection is
established for text and code, where Genetic Prompt treats textual attributes
as genes~\cite{geneticprompt2025} and Genetic-Instruct evolves coding
instructions from a small seed~\cite{geneticinstruct2025}, but in every such
system the genome is linguistic. Engine~C applies the paradigm to a graph
genome that compiles to STIX, which to our knowledge is the first use of
language-model-driven evolutionary operators on structured threat-intelligence
knowledge graphs.

\section{Discriminator Protocol and Timing}
\label{ssec:discriminator-supp}

This section expands the discriminator and control protocol summarized in the
main paper (Section~VII-C). The discriminator is a random-forest
classifier (scikit-learn defaults: $100$ trees, no maximum depth, Gini
impurity) trained on nine graph-level features: edge density, mean degree,
connected-component count, degree skewness, degree Gini coefficient, triangle,
star, and chain motif counts, and relationship-type entropy. Each comparison
uses $5$-fold stratified cross-validation, reporting mean AUC and standard
deviation across folds against a chance baseline of $0.5$. With $N=50$ per
class an AUC of $1.0$ is in principle achievable by overfitting; the
cross-validation protocol and the ER control are jointly the safeguards. Where
Findings~1 and~2 report $\mathrm{AUC}=1.000\pm0.000$, the zero standard
deviation means every held-out fold achieved a perfect score, so we complement
it with the maximum mean discrepancy (MMD), which grades separation rather than
saturating. We use two ER controls: a size-unmatched control for the AUC sanity
check, and a size-matched control drawn from the empirical wargaming size
distribution for the MMD analysis, the stronger null on which the non-random
structure claim rests. Mean wall-clock time per graph was $11.7$--$77.6$\,s
depending on engine (1 to 60 API calls), and the full $N=50$ run took about two
hours.

\section{Downstream Transfer and Augmentation: Full Results}
\label{ssec:downstream-supp}

Table~\ref{tab:kge-inductive} reports the inductive entity-level transfer
experiment of the main paper (Finding~3) in full. The remainder of this
section documents the augmentation experiment of Finding~4.

\begin{table}[!t]
\caption{Inductive entity-level transfer (NodePiece, PyKEEN 1.11.1):
bundle-level entity split, filtered MRR over three seeds. Chance is the
entity-level random-ranker baseline $H_N/N$ for an inference graph of $N$
entities; the Mirage-to-MITRE row is a stress test, degenerate as
anticipated, and the transductive floor is the UUID-level TransE result.
Full configuration in Table~S-II. Relocated from the main paper for length.}
\label{tab:kge-inductive}
\centering\small
\setlength{\tabcolsep}{3pt}
\begin{tabular}{@{}lcccc@{}}
\toprule
Corpus (inference graph) & MRR & Hits@1 & Chance & $N$ \\
\midrule
Mirage $\to$ held-out Mirage & $0.036\pm0.034$ & $0.005$ & $0.029$ & $208$ \\
ER control $\to$ held-out ER & $0.095$ & $0.029$ & $0.038$ & $145$ \\
Mirage $\to$ MITRE (stress) & $0.011$ & $0.002$ & $0.005$ & $1780$ \\
\midrule
Transductive floor (TransE) & $0.019$ & $0.000$ & -- & -- \\
\bottomrule
\end{tabular}
\end{table}

\paragraph{Augmentation configuration} From the MITRE ATT\&CK enterprise
collection we extract $20{,}169$ entity-level triples over six object types
(attack-pattern, campaign, course-of-\allowbreak action, intrusion-set,
malware, tool)
and the four relation types with eligible triples: \textsf{uses}
($18{,}220$), \textsf{mitigates} ($1{,}448$), \textsf{subtechnique-of}
($475$), \textsf{attributed-to} ($26$). The split is $80/10/10$
train/validation/test, stratified by relation type, split seed $42$
($16{,}139$/$2{,}015$/$2{,}015$). All four conditions train TransE and
RotatE with PyKEEN 1.11.1 at embedding dimension $128$, batch size $512$,
Adam at learning rate $10^{-3}$, up to $300$ epochs with early stopping
(frequency $10$, patience $20$), over seeds $11$, $22$, $33$, $44$, $55$
($40$ runs, $0$ failures). Augmentation corpora hold $2{,}117$ triples
each; entities are namespaced (\texttt{AUGM:}, \texttt{AUGER:},
\texttt{AUGSF:}) so they cannot collide with real identifiers, validation
and test contains only real triples, and ranking in every condition is
restricted to the $1{,}792$ real entity labels, preventing the augmented
conditions from being penalized by a larger candidate vocabulary. The
entity-level random-ranker MRR baseline is $H_N/N \approx 0.0042$.
Per-run raw data is archived in
\texttt{output/downstream\_\allowbreak augmentation/\allowbreak results.csv};
the wargaming
scaling raw data referenced in the main paper is in
\texttt{output/wargaming\_scaling\_powered.csv}.

\begin{table}[!t]
\caption{Downstream augmentation, full filtered metrics (mean $\pm$ std
over $5$ seeds) on $2{,}015$ held-out real ATT\&CK test triples.}
\label{tab:augment-full}
\centering
\footnotesize
\setlength{\tabcolsep}{3pt}
\begin{tabular}{@{}llccc@{}}
\toprule
Model & Condition & MRR & Hits@1 & Hits@10 \\
\midrule
TransE & A: real only & $.0514{\pm}.0022$ & $.0136{\pm}.0016$ & $.1169{\pm}.0032$ \\
TransE & B: $+$Mirage & $.0500{\pm}.0040$ & $.0149{\pm}.0030$ & $.1116{\pm}.0069$ \\
TransE & C: $+$ER & $.0478{\pm}.0026$ & $.0126{\pm}.0018$ & $.1094{\pm}.0049$ \\
TransE & D: $+$Schema-F. & $.0492{\pm}.0009$ & $.0126{\pm}.0043$ & $.1123{\pm}.0080$ \\
\midrule
RotatE & A: real only & $.1132{\pm}.0028$ & $.0505{\pm}.0036$ & $.2368{\pm}.0017$ \\
RotatE & B: $+$Mirage & $.1239{\pm}.0033$ & $.0581{\pm}.0029$ & $.2542{\pm}.0030$ \\
RotatE & C: $+$ER & $.1214{\pm}.0019$ & $.0548{\pm}.0016$ & $.2593{\pm}.0050$ \\
RotatE & D: $+$Schema-F. & $.1272{\pm}.0027$ & $.0599{\pm}.0025$ & $.2647{\pm}.0056$ \\
\bottomrule
\end{tabular}
\end{table}

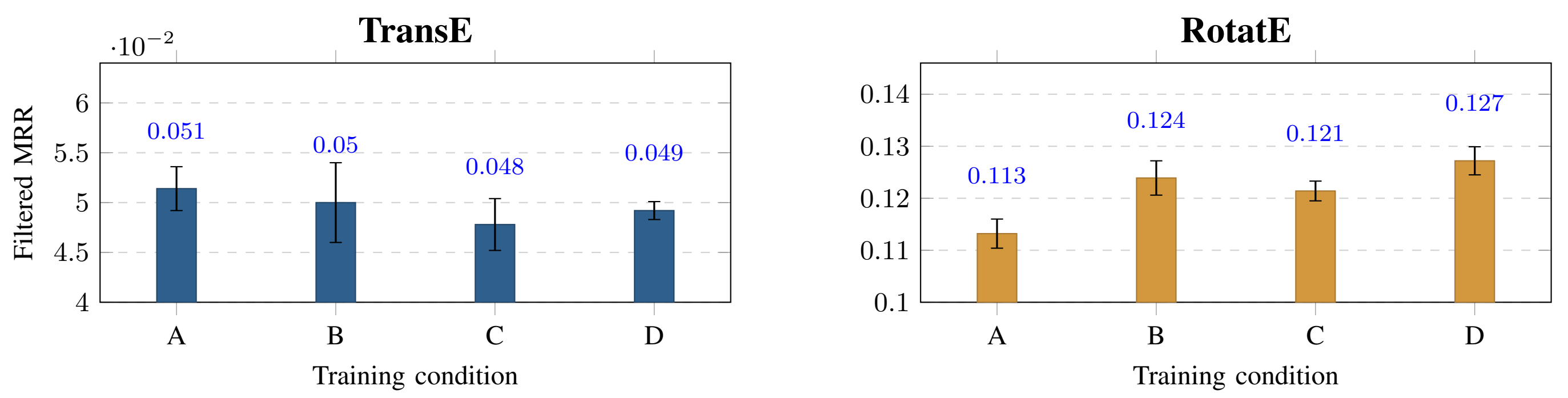
\begin{figure*}[!t]
\centering
\begin{tikzpicture}
\begin{groupplot}[
    group style={group size=2 by 1, horizontal sep=2.2cm},
    width=0.49\textwidth, height=0.24\textwidth,
    ybar=2pt, /pgf/bar width=13pt,
    symbolic x coords={A, B, C, D},
    xtick=data, enlarge x limits=0.16,
    xlabel={Training condition},
    tick label style={font=\small},
    label style={font=\small},
    every axis title/.style={at={(0.5,1.04)}, anchor=south, font=\large\bfseries},
    major grid style={dashed, gray!35}, ymajorgrids,
    error bars/y dir=both, error bars/y explicit,
    error bars/error bar style={black, line width=0.6pt},
    nodes near coords, nodes near coords style={font=\footnotesize, /pgf/number format/fixed, /pgf/number format/precision=3},
    every node near coord/.append style={anchor=south, yshift=13pt},
]
\nextgroupplot[title={TransE}, ylabel={Filtered MRR},
               ymin=0.040, ymax=0.064, ytick={0.040,0.045,0.050,0.055,0.060}]
\addplot+[draw=mblue!75!black, fill=mblue] coordinates {
  (A,0.0514)+-(0,0.0022) (B,0.0500)+-(0,0.0040)
  (C,0.0478)+-(0,0.0026) (D,0.0492)+-(0,0.0009)};
\nextgroupplot[title={RotatE}, ymin=0.100, ymax=0.146,
               ytick={0.100,0.110,0.120,0.130,0.140}]
\addplot+[draw=morange!80!black, fill=morange] coordinates {
  (A,0.1132)+-(0,0.0028) (B,0.1239)+-(0,0.0033)
  (C,0.1214)+-(0,0.0019) (D,0.1272)+-(0,0.0027)};
\end{groupplot}
\end{tikzpicture}
\caption{Downstream augmentation: filtered MRR on held-out real ATT\&CK
triples (mean $\pm$ std over five seeds). A: real only; B: $+$Mirage;
C: $+$ER random control; D: $+$schema-forced. Under RotatE every augmentation
source (B, C, D) clears real-only (A) on every seed, \emph{including the
random control C}, and the three are statistically indistinguishable; under
TransE none helps. The gain is real but structure-agnostic. Full metrics and
Wilcoxon tests in the supplementary material.}
\label{sfig:augment}
\end{figure*}

\begin{table}[!t]
\caption{Wilcoxon signed-rank tests on MRR, paired by seed ($n=5$). At
$n=5$ the one-sided floor is $p=0.0312$ and the two-sided floor is
$p=0.0625$; ``direction'' lists the sign of the per-seed difference in
seed order $11/22/33/44/55$.}
\label{tab:augment-wilcoxon}
\centering
\footnotesize
\setlength{\tabcolsep}{4pt}
\begin{tabular}{@{}llccc@{}}
\toprule
Model & Comparison & Mean diff. & Direction & $p$ \\
\midrule
TransE & B $>$ A (one-sided) & $-.0014$ & $-\,-\,+\,-\,+$ & $0.688$ \\
TransE & C $>$ A (one-sided) & $-.0036$ & $-\,-\,-\,-\,-$ & $1.000$ \\
TransE & D $>$ A (one-sided) & $-.0022$ & $-\,+\,-\,-\,+$ & $0.938$ \\
RotatE & B $>$ A (one-sided) & $+.0107$ & $+\,+\,+\,+\,+$ & $0.031$ \\
RotatE & C $>$ A (one-sided) & $+.0082$ & $+\,+\,+\,+\,+$ & $0.031$ \\
RotatE & D $>$ A (one-sided) & $+.0140$ & $+\,+\,+\,+\,+$ & $0.031$ \\
\midrule
RotatE & B vs.\ C (two-sided) & $+.0025$ & $+\,+\,+\,+\,+$ & $0.062$ \\
RotatE & B vs.\ D (two-sided) & $-.0032$ & $+\,-\,+\,-\,-$ & $0.312$ \\
RotatE & D vs.\ C (two-sided) & $+.0058$ & $+\,+\,-\,+\,+$ & $0.125$ \\
\bottomrule
\end{tabular}
\end{table}

\begin{table}[!t]
\caption{Augmentation corpus statistics and relation-vocabulary overlap
with the real ATT\&CK test vocabulary \{\textsf{uses},
\textsf{mitigates}, \textsf{subtechnique-of}, \textsf{attributed-to}\}.
Entity overlap is zero by construction (namespacing). The gain ordering of
Table~\ref{tab:augment-full} (D $>$ B $>$ C on RotatE MRR) is inverted
relative to the shared-relation fraction, ruling out a knowledge-transfer
channel.}
\label{tab:augment-overlap}
\centering
\footnotesize
\setlength{\tabcolsep}{3pt}
\begin{tabular}{@{}lcccc@{}}
\toprule
Corpus & Rel.\ types & Shared types & Triples on shared & Fraction \\
\midrule
Mirage (B) & 6 & 2 & $936/2117$ & $44.2\%$ \\
ER control (C) & 4 & 4 & $2117/2117$ & $100.0\%$ \\
Schema-forced (D) & 10 & 2 & $530/2117$ & $25.0\%$ \\
\bottomrule
\end{tabular}
\end{table}

\section{Independent Two-Judge Evaluation: Protocol and Full Results}
\label{ssec:judge-supp}

\paragraph{Protocol} Each released bundle is rendered as a plain-text graph summary: one line per node, \texttt{[type] name}, sorted, followed by one line per edge, \texttt{source -verb-> target}, sorted, capped at $120$ total lines with excess edges replaced by an ``...and $K$ more edges'' marker. The summary is scored by two judges from different providers, \texttt{gpt-4o-2024-08-06} and \texttt{claude-sonnet-4-5}, at temperature $0$ with three repeats per bundle, blind to the generating engine, the rubric, and the study design. Main corpus: $200$ bundles $\times$ $3$ repeats $\times$ $2$ judges $=1{,}200$ calls, $0$ failures; classic
generators: $150$ bundles $\times$ $3$ repeats (GPT-4o) $=450$ calls, $0$ failures. The verbatim judge system prompt:

\begin{quote}\small\ttfamily
You are a senior cyber threat intelligence analyst. You will be shown a description of a threat intelligence knowledge graph. Score it on three scales from 1 (very poor) to 5 (excellent): (1) PLAUSIBILITY: Are the entities and their relationships plausible for the cyber threat domain? (2) COHERENCE: Does the graph read as one connected, causally sensible account rather than disconnected fragments? (3) REALISM: Could this plausibly be the structured record of a real intrusion or campaign? Respond with strict JSON: \{"plausibility": n, "coherence": n, "realism": n, "justification": "one sentence"\}. No other text.
\end{quote}

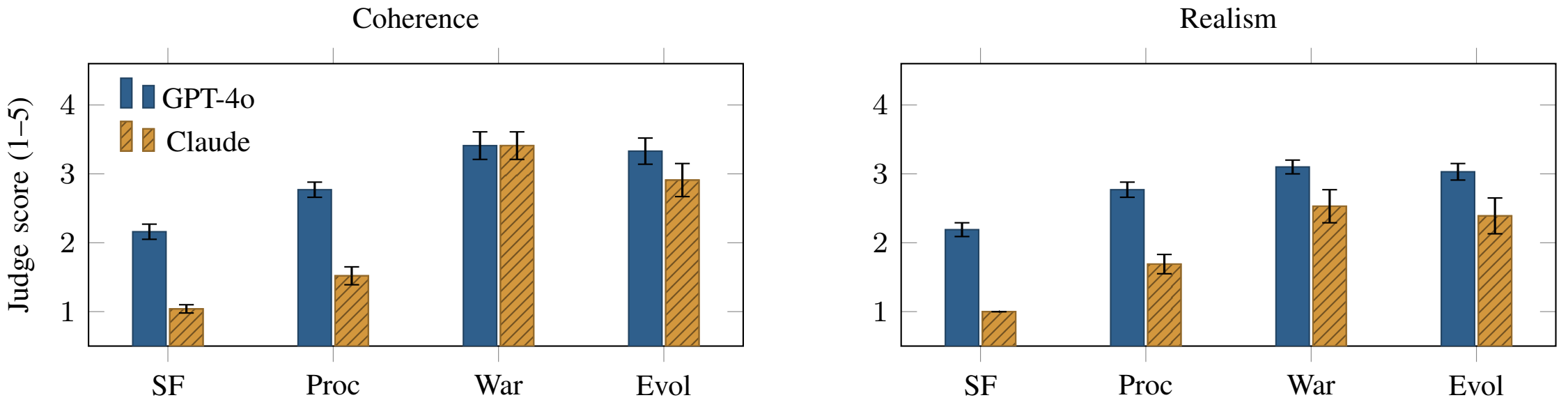
\begin{figure*}[!t]
\centering
\begin{tikzpicture}
\begin{groupplot}[
    group style={group size=2 by 1, horizontal sep=1.5cm, ylabels at=edge left},
    width=0.43\textwidth, height=0.235\textwidth,
    ymin=0.5, ymax=4.6,
    ylabel={Judge score (1--5)},
    ylabel near ticks,
    symbolic x coords={SF, Proc, War, Evol},
    xtick=data,
    tick label style={font=\footnotesize},
    label style={font=\footnotesize},
    title style={font=\footnotesize},
    enlarge x limits=0.16,
    error bars/y dir=both, error bars/y explicit,
    error bars/error bar style={line width=0.6pt},
    legend style={font=\footnotesize, draw=none, fill=none,
                  at={(0.03,0.98)}, anchor=north west},
    legend image code/.code={\draw[#1] (0cm,-0.07cm) rectangle (0.22cm,0.14cm);},
]
\nextgroupplot[title={Coherence}, ybar=1pt, bar width=9pt]
\addplot+[mblue, fill=mblue, draw=mblue!70!black,
    error bars/y dir=both, error bars/y explicit,
    error bars/error bar style={black}]
  coordinates {
    (SF,2.16)  +- (0.11,0.11)
    (Proc,2.77) +- (0.12,0.11)
    (War,3.41) +- (0.18,0.20)
    (Evol,3.33) +- (0.19,0.19)};
\addlegendentry{GPT-4o}
\addplot+[morange, fill=morange, draw=morange!70!black,
    postaction={pattern=north east lines, pattern color=morange!55!black},
    error bars/y dir=both, error bars/y explicit,
    error bars/error bar style={black}]
  coordinates {
    (SF,1.04)  +- (0.04,0.06)
    (Proc,1.52) +- (0.13,0.13)
    (War,3.41) +- (0.19,0.20)
    (Evol,2.91) +- (0.24,0.24)};
\addlegendentry{Claude}
\nextgroupplot[title={Realism}, ybar=1pt, bar width=9pt]
\addplot+[mblue, fill=mblue, draw=mblue!70!black,
    error bars/y dir=both, error bars/y explicit,
    error bars/error bar style={black}]
  coordinates {
    (SF,2.19)  +- (0.09,0.10)
    (Proc,2.77) +- (0.12,0.11)
    (War,3.10) +- (0.09,0.10)
    (Evol,3.03) +- (0.14,0.12)};
\addplot+[morange, fill=morange, draw=morange!70!black,
    postaction={pattern=north east lines, pattern color=morange!55!black},
    error bars/y dir=both, error bars/y explicit,
    error bars/error bar style={black}]
  coordinates {
    (SF,1.00)  +- (0.00,0.00)
    (Proc,1.69) +- (0.15,0.14)
    (War,2.53) +- (0.22,0.24)
    (Evol,2.39) +- (0.27,0.26)};
\end{groupplot}
\end{tikzpicture}
\caption{Two-judge evaluation, bundle means with $95\%$ bootstrap CIs
($N=50$ per engine; full tables in the supplementary material). Both
judges, blind to sources, reproduce the engine ordering; the stricter
Claude judge widens the causal-engine separation and floors schema-forced
realism.}
\label{sfig:judge}
\end{figure*}

\begin{table}[!t]
\caption{Two-judge evaluation, full results: bundle-mean scores with $95\%$ bootstrap CIs, $N=50$ bundles per engine, three repeats per bundle. Kruskal--Wallis across engines is significant on every scale for both judges ($p<10^{-3}$); GPT-4o $H=90.4$ (plausibility), $99.0$ (coherence), $95.3$ (realism).}
\label{tab:judge-full}
\centering
\scriptsize
\setlength{\tabcolsep}{2.2pt}
\begin{tabular}{@{}llccc@{}}
\toprule
Judge & Engine & Plausibility & Coherence & Realism \\
\midrule
GPT-4o & Schema-Forced & $3.24$ [$3.14,3.35$] & $2.16$ [$2.05,2.27$] & $2.19$ [$2.10,2.29$] \\
GPT-4o & Procedural & $3.77$ [$3.65,3.88$] & $2.77$ [$2.65,2.88$] & $2.77$ [$2.65,2.88$] \\
GPT-4o & Wargaming & $4.02$ [$3.96,4.08$] & $3.41$ [$3.23,3.61$] & $3.10$ [$3.01,3.20$] \\
GPT-4o & Evolutionary & $4.01$ [$3.90,4.13$] & $3.33$ [$3.14,3.52$] & $3.03$ [$2.89,3.15$] \\
\midrule
Claude & Schema-Forced & $2.00$ [$2.00,2.00$] & $1.04$ [$1.00,1.10$] & $1.00$ [$1.00,1.00$] \\
Claude & Procedural & $2.43$ [$2.27,2.61$] & $1.52$ [$1.39,1.65$] & $1.69$ [$1.54,1.83$] \\
Claude & Wargaming & $3.82$ [$3.62,4.00$] & $3.41$ [$3.22,3.61$] & $2.53$ [$2.31,2.77$] \\
Claude & Evolutionary & $3.57$ [$3.31,3.82$] & $2.91$ [$2.67,3.15$] & $2.39$ [$2.12,2.65$] \\
\midrule
GPT-4o & Barab\'asi--Albert & $3.81$ [$3.67,3.93$] & $2.89$ [$2.75,3.02$] & $2.80$ [$2.67,2.92$] \\
GPT-4o & Configuration & $3.69$ [$3.54,3.83$] & $2.69$ [$2.53,2.85$] & $2.68$ [$2.52,2.82$] \\
GPT-4o & Watts--Strogatz & $3.82$ [$3.70,3.92$] & $2.91$ [$2.77,3.03$] & $2.81$ [$2.70,2.91$] \\
\bottomrule
\end{tabular}
\end{table}

\begin{table}[!t]
\caption{Judge agreement and stability. Cross-judge agreement is the Spearman correlation of bundle-mean scores between the two judges over all $N=200$ main-corpus bundles ($p<10^{-4}$ on every scale). Repeat stability is the mean per-bundle standard deviation across the three repeats. The judge-versus-rubric correlation uses the mean of realism and coherence per bundle against the rule-based semantic score; within the two LLM baselines, where the rubric is off its ceiling, the correlation is negative (the vacuous-pass artifact of the main paper, Section~VIII-C).}
\label{tab:judge-agreement}
\centering
\footnotesize
\setlength{\tabcolsep}{4pt}
\begin{tabular}{@{}lccc@{}}
\toprule
 & Plausibility & Coherence & Realism \\
\midrule
Cross-judge Spearman $\rho$ & $0.628$ & $0.745$ & $0.680$ \\
Repeat stability, GPT-4o & $0.021$ & $0.026$ & $0.040$ \\
Repeat stability, Claude & $0.009$ & $0.019$ & $0.024$ \\
\midrule
\multicolumn{4}{@{}l@{}}{Judge vs.\ rubric (judge $=$ mean of coherence, realism):} \\
\multicolumn{4}{@{}l@{}}{\quad all $200$ bundles: $\rho=0.450$ ($p<10^{-4}$)} \\
\multicolumn{4}{@{}l@{}}{\quad baselines only ($N=100$): $\rho=-0.375$ ($p=10^{-4}$)} \\
\bottomrule
\end{tabular}
\end{table}

\section{Per-Rule Decomposition with Vacuous Passes, All Generators}
\label{ssec:vacuous-supp}

Table~\ref{tab:vacuous} decomposes the per-rule semantic pass rates of all seven generators into total and \emph{vacuous} passes, where a pass is vacuous if the bundle contains no node of the rule's subject type. The earned score (total minus vacuous, averaged over the eight rules) is the quantity quoted in the main paper (Section~VIII-A). Unlike Table~\ref{stab:perrule}, whose Schema-Forced and Procedural rates come from a dedicated regeneration, these rates are computed over the released bundle set (which, for the wargaming engine, includes regenerated bundles), so values for the four original engines may differ marginally from Table~\ref{stab:perrule}.

\begin{table}[!t]
\caption{Per-rule pass rates, formatted total\,/\,vacuous, over the released bundles ($N=50$ per generator; $N=48$ for Schema-Forced). Leading zeros are dropped. RULE-06 through RULE-08 (vocabulary and integrity) pass non-vacuously everywhere and are summarized in one row. The final row is the earned semantic score: the rule mean of total minus vacuous. Engines: Schema-Forced (SF), Procedural (Pr), Wargaming (Wa), Evolutionary (Ev), Barab\'asi--Albert (BA), Configuration (Cf), Watts--Strogatz (WS).}
\label{tab:vacuous}
\centering
\footnotesize
\setlength{\tabcolsep}{2.6pt}
\begin{tabular}{@{}lccccccc@{}}
\toprule
Rule & SF & Pr & Wa & Ev & BA & Cf & WS \\
\midrule
R-01 & $.62/.00$ & $.06/.02$ & $.96/.00$ & $1./.00$ & $.68/.50$ & $.50/.36$ & $.54/.30$ \\
R-02 & $.16/.00$ & $.90/.88$ & $1./.04$ & $.98/.24$ & $.50/.34$ & $.38/.30$ & $.64/.42$ \\
R-03 & $1./1.$ & $.10/.04$ & $1./1.$ & $1./1.$ & $.52/.38$ & $.44/.38$ & $.48/.44$ \\
R-04 & $1./1.$ & $.00/.00$ & $1./1.$ & $1./1.$ & $.28/.26$ & $.38/.38$ & $.20/.20$ \\
R-05 & $1./1.$ & $.86/.04$ & $1./.04$ & $1./.26$ & $.80/.46$ & $.64/.28$ & $.92/.40$ \\
R-06/8 & $1./.00$ & $1./.00$ & $1./.00$ & $1./.00$ & $1./.00$ & $1./.00$ & $1./.00$ \\
\midrule
Earned & $.47$ & $.49$ & $.73$ & $.69$ & $.48$ & $.46$ & $.50$ \\
\bottomrule
\end{tabular}
\end{table}


\section{Extended Dual-Use Considerations}
\label{ssec:dualuse}
This section expands the dual-use note in the main paper (Section~IX). Mirage
generates synthetic intelligence graphs, typed records of simulated
engagements, rather than executable exploits or operational tooling, and its
world model abstracts services, vulnerabilities, and alerts rather than any
target-specific attack plan. The generator can express attack-pattern
structures absent from published reporting; we judge the incremental offensive
risk modest relative to public knowledge bases such as ATT\&CK, given release
without operational network configurations, and recommend treating generated
graphs as training artifacts rather than validated intelligence. Because the
graphs record simulated rather than observed activity, they carry no live
indicators, host identifiers, or exploit code, and their release exposes no
operational infrastructure.

\section{Language-Model Prompts}\label{ssec:prompts}

For reproducibility, the exact prompts used for every language-model role are preserved verbatim in the released repository (see the main paper, Section~VI, for the repository URL). Five prompts are used, each issued to the policy model with the parameters of the main paper (Section~VI): (i) the wargaming attacker (RED) agent, which receives the attacker's fog-of-war observation and returns a scan or exploit intent, the success of which is decided by the referee rather than by the agent; (ii) the wargaming defender (BLUE) agent, which returns an isolate, deploy, or idle intent; (iii) the procedural semantic-painting prompt, which receives a typed node skeleton and returns names and descriptions only, never edges; and the evolutionary (iv) mutation and (v) crossover operators, which receive one or two parent genomes in the IR and propose a single offspring while selection remains deterministic. In every case, the prompt constrains the model to emit only IR or JSON content and never to assert relationship success, which is the textual counterpart of the architectural separation described in the main paper (Section~IV-A). A sixth prompt, the two-judge evaluation prompt, is an evaluation role rather than a generation role and is reproduced verbatim in Section~\ref{ssec:judge-supp}.

\end{document}